# A Flexible Method for Behaviorally Measuring Alignment Between Human and Artificial Intelligence Using Representational Similarity Analysis

**Mattson Ogg[1], Ritwik Bose, Jamie Scharf, Christopher Ratto, Michael Wolmetz**

*Research and Exploratory Development Department*

*Johns Hopkins University Applied Physics Laboratory, Laurel, Maryland, USA*

mattson.ogg@jhuapl.edu; rik.bose@jhuapl.edu; jamie.scharf@jhuapl.edu; christopher.ratto@jhuapl.edu; michael.wolmetz@jhuapl.edu

## Abstract

**As we consider entrusting Large Language Models (LLMs) with key societal and decision-making roles, measuring their alignment with human cognition becomes critical. This requires methods that can assess how these systems represent information and facilitate comparisons with human understanding across diverse tasks. To meet this need, we adapted Representational Similarity Analysis (RSA), a method that uses pairwise similarity ratings to quantify alignment between AIs and humans. We tested this approach on semantic alignment across text and image modalities, measuring how different Large Language and Vision Language Model (LLM and VLM) similarity judgments aligned with human responses at both group and individual levels. GPT-4o showed the strongest alignment with human performance among the models we tested, particularly when leveraging its text processing capabilities rather than image processing, regardless of the input modality. However, no model we studied adequately captured the inter-individual variability observed among human participants, and only moderately aligned with any individual human's responses. This method helped uncover certain hyperparameters and prompts that could steer model behavior to have more or less human-like qualities at an inter-individual or group level. Pairwise ratings and RSA enable the efficient and flexible quantification of human-AI alignment, which complements existing accuracy-based benchmark tasks. We demonstrate the utility of this approach across multiple modalities (words, sentences, images) for understanding how LLMs encode knowledge and for examining representational alignment with human cognition.**

## Introduction

Foundation Model (FM) reasoning (Brown et al., 2020; Wang et al., 2024) and perceptual skills (Marjieh et al., 2024; Radford et al., 2021) may soon match or exceed human performance across a wide range of tasks (OpenAI et al., 2024; Wei et al., 2022; see Minaee et al., 2024 for review). The rapid pace of this progress, as exemplified by Large Language Models (LLMs) (Hoffmann et al., 2022; Kaplan et al., 2020), has initiated a discussion of whether and how these models should be integrated into every-day life or be given additional responsibilities (Amodei, 2024). The increasing deployment of AI systems in critical roles (potentially replacing humans in those roles) requires scalable, generalizable methods for measuring how FMs represent knowledge about the world, and for evaluating how those representations and downstream behaviors compare to complex human behaviors (see Sucholutsky et al., 2023 for discussion). The recent transition from fundamentally narrow models of moderate size and complexity to increasingly general models that are larger and more complex compounds the classic opacity problem of deep neural networks: discerning how an LLM processes a given input or arrives at a decision has never been as challenging or as critical. Fortunately, cognitive science and psychological research has focused on this exact challenge in the context of biological intelligence and serves as a productive framework for studying artificial cognition and its alignment with human cognition.

[1] Correspondence can be addressed to Mattson Ogg, Research and Exploratory Development Department, Johns Hopkins University Applied Physics Laboratory, Laurel, Maryland, USA, mattson.ogg@jhuapl.edu

As an exemplar of this emerging framework, *Turing Experiments* capitalize on the rich history of experimental psychology to measure the cognitive and behavioral alignment between human and artificial intelligence (Aher et al., 2023; Mei et al., 2024). In a Turing Experiment, an LLM is used to simulate a sample of the human population over repeated runs (Cava & Tagarelli, 2024; Mei et al., 2024), sometimes using simulated participant identities for each new run (Aher et al., 2023; Petrov et al., 2024), and is prompted to engage in classic psychology tasks (e.g., the prisoner's dilemma, the ultimatum game, or the Milgram shock experiment). While still very early, this line of work has provided insight into LLM reasoning and has highlighted similarities with human behavior. For example, the GPT family of models (as well as different open-source models) can exhibit discrete personalities based on their prompting (Cava & Tagarelli, 2024) that influence behavior (Bose et al., 2024). These models can also generate responses in behavioral tasks that fall within the range of human variability (Mei et al., 2024), and, finally, larger, more recently developed language models better align with human behavior (Aher et al., 2023). This approach allows for an examination both of how similar LLM behavior is to human behavior in the aggregate (i.e., using the central tendency of a sample of simulated participants) but can also help us understand inter-individual variability among human and AI participants. These initial studies suggest there could be value in using other paradigms and techniques to explore the knowledge and behavior of LLMs and how they align with human knowledge and behavior. There is an active discussion regarding how best to use this framework to support psychological research and how, if executed properly, LLMs can help further our understanding of human cognition (Abdurahman et al., 2024; D'Alessandro & Thompson, 2025). We note, however, that here we are interested in using behavioral testing paradigms borrowed from cognitive science to improve our understanding of the cognition of LLMs (Ivanova, 2025), rather than humans.

One of the most productive methods for mapping the structure of how an individual represents information about the world is the use of pairwise ratings of similarity (or dissimilarity) with respect to a pair of stimuli (Shepard, 1980, 1987; Tversky, 1977). This class of tasks is adaptable to a wide array of domains and questions (e.g., "How similar are the words 'apple' and 'hand?'" or "How similar are these two images?"), and is most useful when the experimenter does not have direct access to a participant's internal representations (i.e., neuronal activations or embeddings), as is the case in traditional psychophysics and cognitive science experiments as well as for many frontier LLMs. Ratings elicited by participants on each trial comprise a behavioral distance metric for the two stimuli that were presented. These ratings can be organized into a symmetrical matrix whose rows and columns correspond to the probe items used in each rating trial and can be analyzed using techniques like multi-dimensional scaling to visualize the geography of how different items relate to one another (Hout et al., 2013) or to test different hypotheses (Borg & Groenen, 1997). This approach has deeply informed a range of questions in human perception and cognition including object relations (Jiang et al., 2022; Ogg & Slevc, 2019) and semantic information (Carlson et al., 2014), as well as musical pitch (Marjieh, Griffiths, et al., 2023) and timbre (McAdams et al., 1995; Thoret et al., 2021).

Representational similarity analysis (RSA; Kriegeskorte & Kievit, 2013) builds on the use of distance, or dissimilarity, matrices ("DSMs" including from pairwise ratings) to quantify the similarity of *representational spaces* among diverse systems: across organisms (Kriegeskorte et al., 2008), individuals (Khaligh-Razavi & Kriegeskorte, 2014), models (Mehrer et al., 2020; Ogg & Skerritt-Davis, 2021), or biological substrates such as different brain regions (Carlson et al., 2014; Giordano et al., 2023; Ogg et al., 2020). In RSA, the organized distance matrices are correlated with one another to quantify the agreement of the pairwise ratings (or distances) between each system. That is, RSA can be used to quantify how similarly two species (e.g., humans and primates; Kriegeskorte et al., 2008) process object images at different stages of the visual hierarchy, to align object representations from different neuroimaging modalities across time and cortical space (Cichy et al., 2014), or to investigate how the computations performed by layers of convolutional networks trained for visual object recognition relate to the computations of the ventral visual pathway (Cichy, Khosla, et al., 2016).

RSA has also been used to understand the representations of neural network models (e.g., (Mehrer et al., 2020, 2021; Ogg, 2025; Ogg & Skerritt-Davis, 2021; Sucholutsky et al., 2023) by correlating distances derived from model embeddings. However, for current LLMs, these embeddings are not always accessible. Instead, the representational structure of these models can be distilled by formulating this analysis as a Turing Experiment where the model is queried with pairs of stimuli and asked to provide a similarity rating for each pair. A crucial advantage of combining pairwise ratings with RSA is that it enables comparison between any systems capable of producing comparable behavioral outputs, without requiring access to or assumptions about their internal representations. This makes it particularly valuable for comparing human and artificial intelligence, where internal processing mechanisms may be fundamentally different or inaccessible.

A growing body of work has begun to adapt pairwise rating methods to behaviorally probe LLMs and to measure their alignment with humans. A series of studies by Marjieh and colleagues (Marjieh et al., 2022; Marjieh, Rijn, et al., 2023; Marjieh et al., 2024) have explored a continuous approach to mapping LLM knowledge via representational distances either from model embeddings or model ratings. Their initial results found that LLMs can predict human similarity judgments across multiple perceptual domains based on text input alone. Dickson and team (2024) asked similar questions regarding perception based on visual input, finding that different LLM models aligned with human ratings along some (but not all) perceptual dimensions. Finally, Du and colleagues (2025) studied LLM and VLM perception via a alightly different leave-one-out oddball task similar to pervious work with human raters (Hebart et al., 2020), finding high alignment with human ratings and neural responses. However, hese analyses were primarily focused on visual object ratings and RSA analyses were focused primarily on model embeddings and neural responses rather than directly measuring pairwise rating behavior. None of these previous studies have compared representations across text and image domains or undertook an evaluation of the variability among ratings of LLM participants.

We build on this prior work at the intersection of Turing Experimentation and RSA by designing a behavioral pairwise rating task to probe the knowledge and behavior of LLM agents as a strategy for measuring alignment between artificial and biological intelligence across information domains (e.g., text and images). Using this method, we measure the relationship between LLM and human judgments for sets of well-studied probe objects via words and images. In the process, we demonstrate the flexibility to facilitate comparisons within and across different modalities (i.e., text and images), quantify individual variability among LLMs and humans, and evaluate prompt effects among standard Turing Experiment formulations.

# Methods

## Language models

We elicited responses from different versions of OpenAI’s Generative Predictive Transformer (GPT) models (Brown et al., 2020; OpenAI et al., 2024) hosted on the Azure cloud computing platform via the API (version “2023-03-15-preview” for GPT-4o models and version "2023-05-15" for all other models). The following GPT models were selected: GPT-3.5 Turbo Model with a 16k context window (“gpt-35-turbo-16k,” model version “0613,” referred to as “GPT-3.5”), a GPT-4 (text-only) model (“gpt-4,” model version “1106-preview,” referred to as “GPT-4”), a GPT-4-Vision model (“gpt-4,” model version “vision-preview,” referred to as “GPT-4-Vision”), and GPT-4o (“gpt-4o,” model version “2024-08-06,” referred to as “GPT-4o”) and GPT-4o-mini model (“gpt-4o-mini,” model version “2024-07-18,” referred to as “GPT-4o-mini”). For each experiment run (i.e., simulated participant) the model was initialized using a specific temperature value, and otherwise used default parameters. Unless stated (i.e., during specific follow up experiments where temperature values were swept over a set: 0.01, 0.7, 1 and 1.5) a temperature value of 1.0 was used throughout (no temperature sweeps for the "GPT-4o" family of models, these models were all run at 1.0). For GPT-4-Vision models, 4096 max tokens were specified.

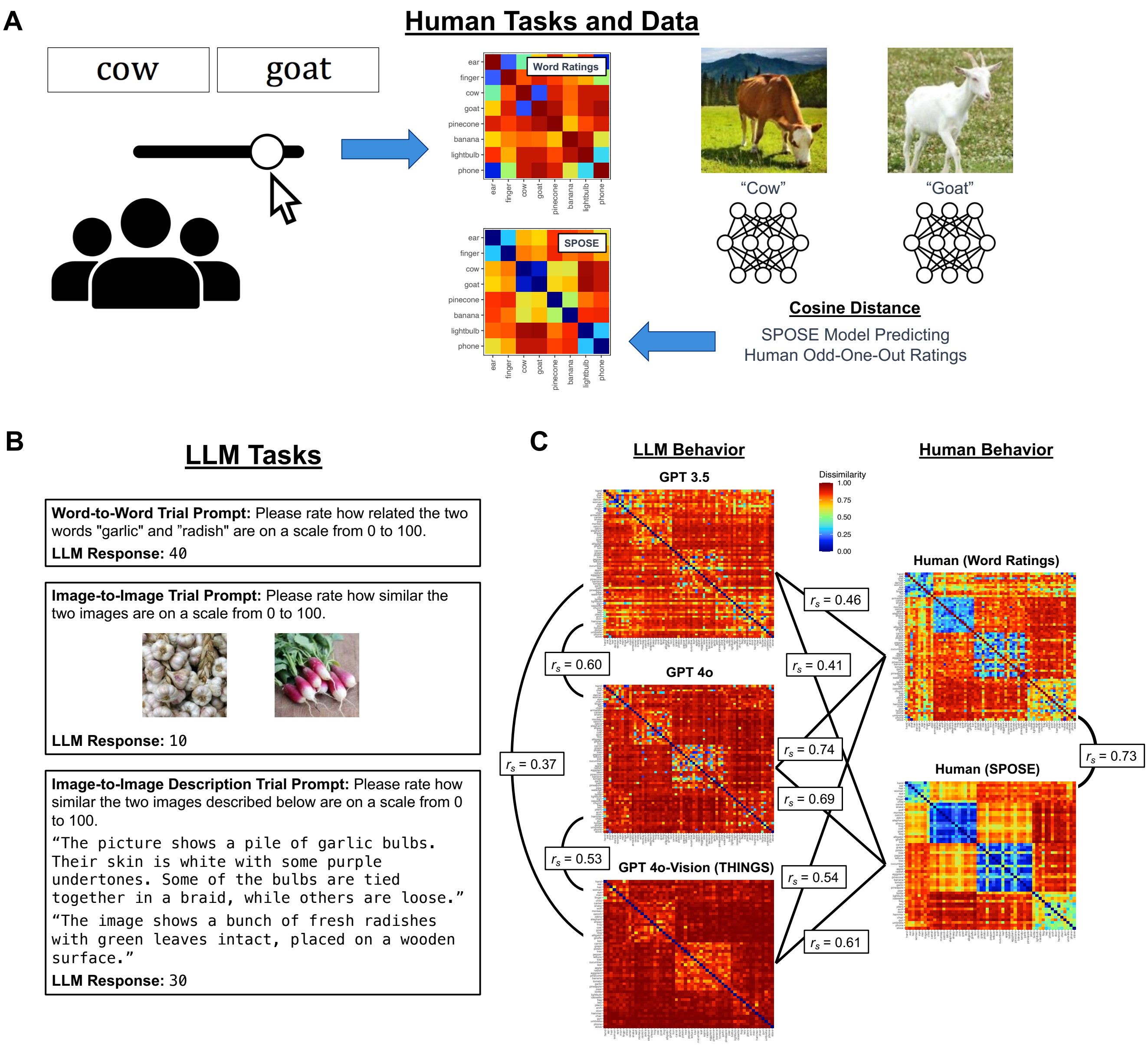

*Figure 1* Summary of the behavioral pairwise rating and RSA methodology and results. **(A)** Depiction of the task and models used to derive pairwise human data via word ratings (Left) and a model of human responses on an odd-one-out task for images, called Sparse POsitive Similarity Embeddings (SPOSE) (Right). **(B)** Synopsis of a pairwise rating behavioral RSA trial for each of our main experiments. LLM responses for each trial are entered into the intersecting cell for the item pair in the corresponding dissimilarity matrices (DSMs). **(C)** DSMs for LLMs ($n$ = 24 in each matrix) and Humans averaged over participants ($n$ = 8 for word ratings). The Spearman rank correlation values between them summarize their representational alignment (all Bonferroni-corrected $p < 0.05$). See text and Supplemental Tables 2 and 3 for additional details as well as Examples 1, 2 and 3 for a depiction of the full prompts for each experiment.

The GPT family of models have been among the most popular and widely used LLM tools and are thus an important class of models to understand. However, this generation of GPT models are closed source and API updates may pose challenges for reproducibility. Thus, we also evaluated a series of open-source models using the Ollama platform.

These quantized models were all run locally on a personal computer (2021 16-in M1 Macbook Pro with 16gb of memory), so in most cases these were run using smaller instantiations (7 to 14 billion parameters) for better throughput. This service provided access to Gemma (7b, 430ed3535049), Gemma-2 (9b, ff02c3702f32), Phi-3-medium (14B, 1e67dff39209), Mistral (7b, 61e88e884507), Solar (10.7b, 059fdabbe6e6), Llama-2 (7b, 78e26419b446), Llama-2-uncensored (7b, 44040b922233), and Llama-3 (8b, 71a106a91016) for our text experiments. These models were also re-initialized for each experiment and run using a temperature of 1.0.

### Baseline neural network models

Baseline text model embeddings were obtained from public sources. From OpenAI, text embeddings were extracted from the Ada model ('text-embedding-ada-002' version 2, referred to as "Ada") using the same software infrastructure as the GPT models above. GloVe (Pennington et al., 2014) embeddings were obtained from an online repository (version: Common Crawl 840B tokens, 2.2M vocab, cased, 300d vectors: https://nlp.stanford.edu/projects/glove). Finally, two BERT variants from https://huggingface.co (Wolf et al., 2020) along with their tokenizers were used: a standard BERT model ("bert-base-uncased;" Devlin et al., 2019) as well as a larger variant ("albert-xxlarge-v2;" Lan et al., 2020) that has been shown to align well with human neural responses (Schrimpf et al., 2021). For each of the baseline text models, we extracted embeddings for each of the 67 words in our text stimulus set and then computed the cosine similarity between the model embeddings for each pair of words.

Two AlexNet models published along with the Ecoset dataset (Mehrer et al., 2021) were used as baselines in our GPT-4-Vision experiments. One variant of these models was trained on the original ImageNet Large Scale Visual Recognition Challenge (ILSVRC) 2012 data (denoted "ILSVRC2012 AlexNet") while the other variant was trained on the Ecoset data (denoted "Ecoset AlexNet"). These specific model architectures achieved the best correspondence with human behavioral data in the experiments conducted by Mehrer and colleagues (2021). Specifically, versions of the model that were trained on the Ecoset data were found to produce image classification models whose embeddings aligned better with human neural and behavioral data than models trained on ILSVRC 2012. Model variants initialized with "training seed 01" were used for each Ecoset and ILSVRC2012 model. Embeddings from layer 7 of both models' responses to each image (rescaled to 224 by 224 resolution) were extracted and cosine distances between these embeddings from each image in the stimulus sets was used to populate the models' DSMs. A gallery of all the DSMs generated for these experiments (averaged at the group level where appropriate) can be found in Supplemental Figure 1.

### Semantic word rating task

We compared human and LLM responses on a task judging the semantic relatedness of words based on the experiments and data from Carlson and colleagues (2014). In this task, participants judged the semantic relatedness of a set of 67 words (see Supplemental Table 1), which correspond to a subset of well-studied image stimuli depicting common objects (Cichy, Pantazis, et al., 2016; Kriegeskorte et al., 2008). From the original data of Carlson and colleagues (2014), we retained data from 8 participants who completed all three sessions of the task. On each trial participants were presented with a pair of words and asked to rate (using a slider) how semantically related the two objects were. The slider position was converted into a value between 0 and 50 that was recorded and analyzed (these values were re-scaled from 0 to 100 for comparability with the LLM outputs). Each participant rated each pair of words (one unique word order for each pair) for a total of 2211 ratings. Note, data from these stimulus pairs were mirrored to fill out the opposite ordered pairs for DSM visualizations, but these mirrored entries were not included in our other analyses.

The task for the LLMs was modelled as closely as possible on the task administered to the human participants, with minor modifications to accommodate model responses, and to minimize errors (see Example 1). After an initial prompt introducing the task (and where applicable, the LLM participant's surname and honorific), each trial prompted the LLM to respond with a number from 0 to 100 to characterize the relatedness of a pair of words (Shah et al., 2023). Following Aher and colleagues' (2023) original Turing Experiment formulation, in some experiments the LLM was assigned to assume a participant identity using a surname and honorific. This information was included at the beginning of each prompt when addressing the model. Surnames (Snyder, Smalls, Rodriguez, Olson, Nguyen, Kim, Jeanbaptiste, Garcia) and honorifics (Ms., Mr., Dr.) were drawn from a representative sample taken from the larger set used by Aher and colleagues (2023), with the addition of the 'Dr.' honorific. The crossing of each surname with each honorific produced a cohort of 24 simulated participants for each of our word-to-word rating experiments. When the experiment was run for the LLM without any participant identifiers, all surnames and honorifics were removed from the prompts thus not invoking any specific identity for that participant's run. LLM participants rated all possible pairs of items (including both orders of unique items) for a total of 4489 trials. The order of the word pairs was shuffled for each participant. An example portion of one of the LLM experiments is provided in Example 1. Note that for some combinations of the Kim, Nguyen and Jeanbaptiste surnames involving the objects pineapple, woman and cow, trials queried to GPT-4o and GPT-4o-mini were flagged by the OpenAI content filter (presumably by mistake). In such cases those trials were skipped, but future work examining differences between models with and without guardrails or content filters would be well suited to follow up work.

## Image similarity rating task

We adapted the Word-to-Word Semantic Similarity Rating Task to accommodate relatedness ratings for images corresponding to each object. We selected two sets of exemplar images for each object (i.e., for each word stimulus) from public databases. One set of stimuli comprised a set of images depicting each object in the real world that was then cropped and presented on a gray background (using the study materials posted online by Cichy, Pantazis, et al., 2016, which are similar to Kriegeskorte et al., 2008, these are referred to as the "Carlson-Image" stimuli since their behavior ratings were used for analyzing neural responses to these images). This is a well-studied stimulus set in the field of cognitive neuroscience but it has numerous shortcomings as discussed in Grootswagers and Robinson (2021). Thus, another set of more natural object images that included representative backgrounds was drawn from the THINGS database (Hebart et al., 2019, 2023), wherein a single exemplar was selected from the set of images within categories corresponding to each object label. Note, matching categories existed for 55 of the 67 objects across the Carlson and THINGS datasets. Thus, all comparisons with models or responses derived from the THINGS dataset were constrained to this subset of 55 objects where THINGS and the original Carlson and colleagues (2014) stimulus sets overlap. Note that a small number of these image stimuli (mostly THINGS stimuli depicting the 'hair' and 'ear' object classes) failed because they tripped the OpenAI content filter (ostensibly by mistake), which required us to curate our THINGS image set to arrive at a stimulus set that was usually successfully processed (Supplemental Table 1 lists the image files used).

All images were resized (to a resolution of 150 by 150), converted to base64 and packaged into a dictionary for presentation to the model via the API. See Example 2 for a depiction of task instructions and an example of how this experiment proceeded for a representative participant. This again resulted in 4489 ratings for the 67 Carlson Images or 3025 ratings for the 55 THINGS images. These analyses were mostly restricted to comparisons among the 55-stimulus set for comparability but results for the full Carlson-Image dataset are reported in Supplemental Table 2. The order of the image pairs was shuffled for each participant. Because the GPT-4-Vision models were more expensive, these were run for a reduced number of simulated participants for the image processing experiments (crossing surnames: Garcia, Nguyen, Olson, and Smalls with honorifics Ms. And Mr.) for a total of 8 simulated participants for each experiment run.

The human data for our Image-to-Image rating experiments comprised behavioral embeddings generated by a large number of leave-one-out ratings for images from the THINGS dataset (see Hebart et al., 2020 for details). This embedding was obtained from a model that was able to accurately reproduce human behavioral judgements on this task (at the noise-ceiling with respect to the human behavioral data). We retained the embedding corresponding to each of the 55 THINGS object classes (that overlapped with the Carlson classes) and computed the cosine distances between them.

We found the DSMs produced by GPT-4-Vision to be very sparse and wondered if this was related to the vision or text processing modules. Indeed, Yuksekgonul and colleagues (2023) suggest that vision-language models can perform poorly on relational understanding and linking tasks. To compare this model's native image processing capabilities to its text processing capabilities, another set of experiments was run where the LLM participant was first asked to provide a description of each image, and then on subsequent trials, the model rated the similarity of the images based solely on the text descriptions it had just provided. This is similar to the approach of Marjieh and colleagues (2022), except we used a single GPT-4 image processing model instantiation for the entire experiment, rather than generating descriptions with one model and comparing text embedding distances using another model. Thus, images were distilled to text descriptions and then ratings were made based only on the text descriptions. Since the GPT-4o models were capable of natively processing both text and images, this allowed for cross modal analyses of that model's specific semantic representations. A summary depiction of this task for a representative participant is provided in Example 3. Note that the “man" and “apple” stimuli were removed from the image description task of the “Carlson-Image” stimuli for the GPT-4o models since they tripped the model's content filter.

## Representational similarity analyses

To analyze alignment between models and humans on these pairwise rating tasks, the responses among the participants for each model (or among humans) were averaged to a group-level DSM followed by a Spearman rank correlation among the flattened, group-level dissimilarity matrices. For completeness, we evaluated all pairwise correlations among model systems at the group level with a stringent Bonferroni correction applied to assess statistical significance (using a 528 and 1225 comparison correction among models that assessed the 67-item Carlson dataset and the 55-item THINGS dataset, respectively, see Supplemental Tables 2 and 3) against a null hypothesis of no correlation between the two DSMs. Group level comparisons were evaluated using two-sample Wilcoxon tests. To evaluate inter-subject agreement, we computed correlations among individual participant's dissimilarity matrices. Two-way intraclass correlation coefficients for agreement over the ratings of each model were also computed using the ‘irr’ package in R. Finally, individual-level human-LLM alignment was assessed via correlation between individual LLM participant DSMs and DSMs from individual human participants.

Throughout these experiments, we observed a small number of non-compliant trials similar to previous work (Hansen & Hebart, 2022; Mei et al., 2024), where the LLM provided a verbose reply (ignoring prompted instructions to reply only with a number), replied that they were an AI agent and not a participant in our study (ignoring the task prompt), or replied that a query triggered a content filter. These accounted for < 1% of the trials obtained for the GPT models and were removed from further analysis. Content filter issues mostly pertained to the GPT-4-Vision models, and we preselected image stimuli that minimized these errors as much as possible. GPT-4o had some issues with pineapple, cow and woman ratings for the Kim, Nguyen and Jeanbaptiste surnames, which were skipped. Finally, some open-source text models had difficulties with certain words (e.g., ‘gun’ or ‘woman’ for Llama-2 or Gemma). Some open-source models also provided verbose or formulaic replies (e.g., Mistral almost always explained its reasoning regardless of our prompts), which were parsed, cleaned or (where necessary) removed from further analysis.

# Results

## Word similarity judgements

We designed a prompt-based task to map the high-level semantic representational space of LLMs, drawing inspiration from the power and ostensible simplicity of pairwise rating tasks (Figure 1A, and e.g., Marjieh et al., 2024). Even without direct access to the model's embeddings or internal representations, the flexibility of the chat prompt interface allowed us to repeatedly query the model with questions asking the LLM to rate the similarity of two concepts (see

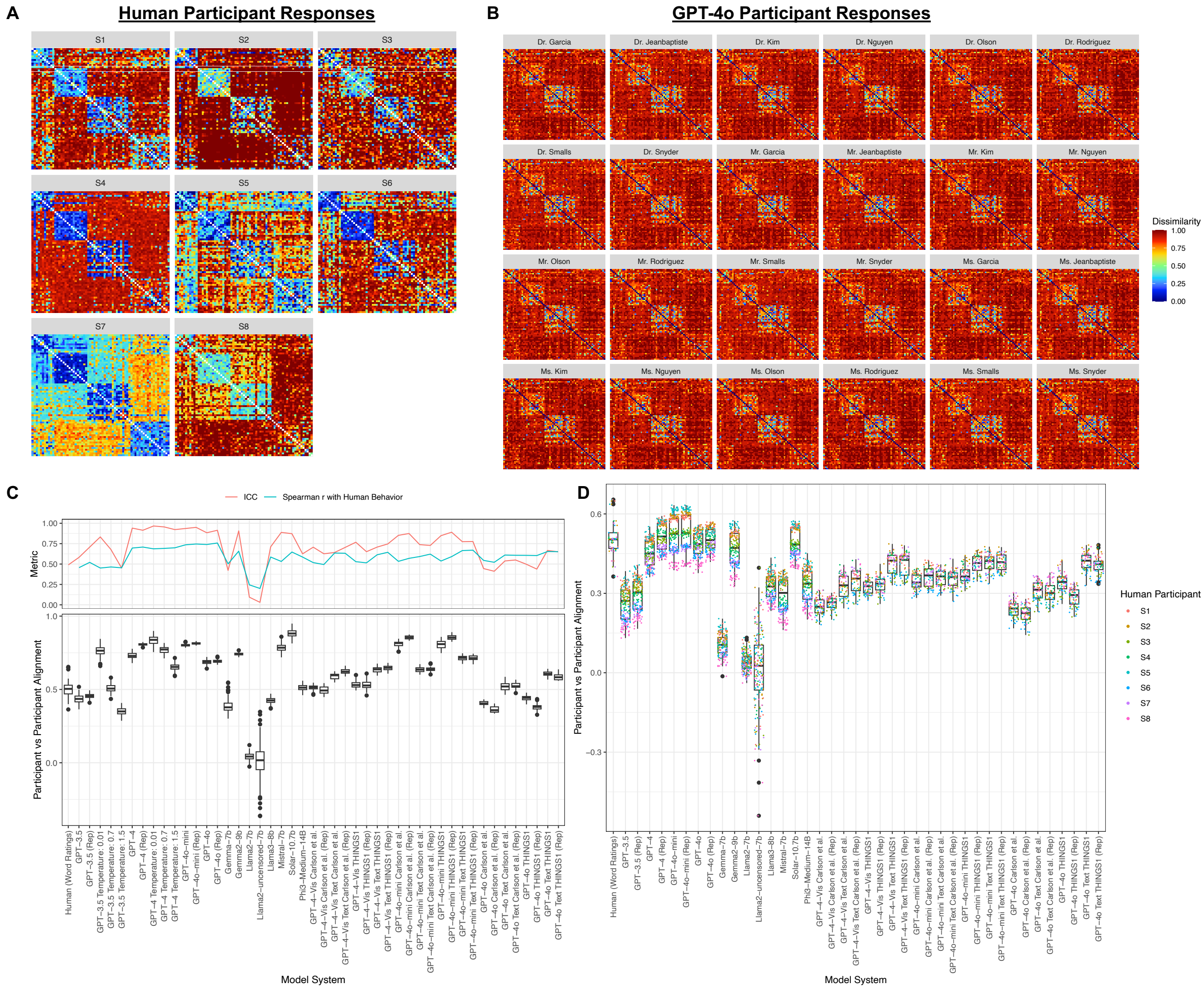

*Figure 2* Quantifying variability among participants for each model system. **(A)** DSMs for each of the human participants who provided word-similarity ratings. **(B)** DSMs for a cohort of GPT-4o participants. **(C)** Summary of variability among individuals of each model system, (top) plotting aggregate ICC values and group-level Spearman correlation statistics across each set of real and simulated participants, and (bottom) a plot of the distribution of inter-subject representational alignment for each model system. Temperature was set to 1.0 unless otherwise stated. **(D)** individual-level cross-system alignment between individual LLM and human participants. Alignment is organized by model system along the x-axis and colored points indicate the human participant in the comparison.

Figure 1B and Example 1). This method also allowed us to quantify the representational similarity (or representational alignment) of two model systems, and to specifically assess how similar LLM semantic representations are to those of humans. Repeated runs for a given model allowed us to examine the variability of LLM responses by assigning the model a different simulated *participant* identity for it to assume on each new run (see Aher et al., 2023; Cava & Tagarelli, 2024; Mei et al., 2024; Petrov et al., 2024). The LLM was assigned these identities (for example, “Ms. Olson”) each time it was initialized for a given participant run: both when introducing the task and during the subsequent run of rating trials (see Example 1).

To validate this approach, we compared LLM responses to previously collected pairwise ratings of semantic relatedness for different object concepts provided by a cohort of human raters. In the original study eight participants judged the semantic relatedness of all pairs of 67 object concepts, originally collected for comparison with neural responses to assess when semantic meaning emerges within the human ventral visual pathway (Carlson et al., 2014, see Figure 1C and Supplemental Table 1). The same task instructions and word stimuli were presented to GPT-3.5, GPT-4, GPT-4o-mini, GPT-4o and other smaller open-source language models to elicit similarity ratings that could be compared to human ratings (Figure 1B).

GPT-4o ratings were most similar to the human participant responses among the models that were tested ($r_s$ = 0.740; all Bonferroni-corrected $p < 0.05$ unless otherwise stated, see Figure 1C and Supplemental Table 2 for more details and for the full set of comparisons, see Supplemental Figure 1 for group-level DSMs), showing much stronger alignment than GPT-3.5 ($r_s$ = 0.456) and GPT-4 ($r_s$ = 0.696) ratings. However, GPT-4o-mini ($r_s$ = 0.736) aligned with human ratings almost as well as GPT-4o. Among smaller language models and language model embeddings, Gemma2-9b ratings ($r_s$ = 0.658), Solar-10.7b ($r_s$ = 0.647), Llama3-8b ($r_s$ = 0.586), Phi-3-medium-14b ($r_s$ = 0.586), Mistral-7b ($r_s$ = 0.532), GloVe embeddings ($r_s$ = 0.643) and Ada embeddings ($r_s$ = 0.437) were closely aligned with human ratings. In many cases these smaller LLMs were more aligned with human ratings than GPT-3.5, but not as closely as GPT-4 ratings. Llama2 (Llama2-7b: $r_s$ = 0.245; Llama2-uncensored-7b: $r_s$ = 0.201) and BERT (bert-base-uncased: $r_s$ = 0.188) models had the lowest alignment with human ratings among the text models that were tested (the albert-xxlarge-v2 alignment of $r_s$ = -0.001 did not survive Bonferroni correction).

The stimuli in these experiments were organized into discrete object categories (Human, Animal, Natural, and Man-Made objects, see Supplemental Table 1), which allowed us to examine how these models represented different aspects of within- and between-category semantic structure. Most models were more aligned with human participant ratings of items within the same object category (GPT-3.5: $r_s$ = 0.472; GPT-4: $r_s$ = 0.780; GPT-4o-mini: $r_s$ = 0.747; GPT-4o: $r_s$ = 0.781; Gemma-7b: $r_s$ = 0.479; Gemma2-9b: $r_s$ = 0.691; Llama3-8b $r_s$ = 0.613; Phi-3-medium-14b $r_s$ = 0.498; Mistral-7b: $r_s$ = 0.482; Solar-10.7b: $r_s$ = 0.715; GloVe: $r_s$ = 0.801; Ada: $r_s$ = 0.660, all Bonferroni-corrected $p < 0.05$), compared to items from different categories (GPT-3.5: $r_s$ = 0.293; GPT-4: $r_s$ = 0.510; GPT-4o-mini: $r_s$ = 0.602; GPT-4o: $r_s$ = 0.598; Gemma-7b: $r_s$ = 0.389; Gemma2-9b: $r_s$ = 0.474; Llama3-8b $r_s$ = 0.377; Phi-3-medium-14b $r_s$ = 0.481; Mistral-7b: $r_s$ = 0.466; Solar-10.7b: $r_s$ = 0.470; GloVe: $r_s$ = 0.485; Ada: $r_s$ = 0.246, all Bonferroni-corrected $p < 0.05$; W = 292, $p < 0.012$, two-sided Wilcoxon Rank-Sum test for within vs between category text model alignment). In general, the models that were better aligned with human participant ratings were most aligned with respect to within-category structure and their between-category structure was very sparse (i.e., high performing models were more aligned when rating objects from the same category such as “cow” and “goat” than when rating objects from different categories like “cow” and “phone” and these between category ratings tended to be near zero; see Supplemental Table 2 and Supplemental Figure 1). This indicates that there is some nuance in how human raters represent some between-category semantic relations that is not well captured by the LLMs.

## Image similarity judgements

Initial experiments demonstrated that our pairwise rating task allowed us to behaviorally probe the structure of an LLM’s semantic representations and compare those with humans. Next, we evaluated the generalizability of this method for comparing human and LLM behavioral judgments across domains. For this we used the images corresponding to the object words rated in the study by Carlson and colleagues (2014; obtained from Cichy, Pantazis, and colleagues, 2016, although they originated from Kriegeskorte and colleagues, 2008) referred to as the “Carlson-Image” stimuli (the human behavior ratings were collected to analyze neural responses to these images). We ran an additional set of experiments using corresponding images from the THINGS database (Hebart et al., 2019; Hebart et al., 2023). This complemented the Carlson-Image stimulus set, which comprises a cropped view of each object presented on a grey background, with more natural depictions of each object (including backgrounds) typical of the THINGS dataset (see Grootswagers and Robinson, 2021, for discussion). There was incomplete overlap in the object classes between these two stimulus sets, so analyses were restricted to models or responses for the 55 object classes present in both the Carlson-Image stimulus set and the THINGS database (see Supplemental Table 1).

We adapted the text-based pairwise rating task to elicit similarity ratings for pairs of images from the GPT-4-Vision and GPT-4o models (see Figure 1B and Example 2). For a comparison with human behavioral ratings, we used cosine distances between Sparse POsitive Similarity Embeddings (SPOSE) generated for each object (Hebart et al., 2020). These embeddings were learned so as to accurately predict odd-one-out behavioral judgments for a large number of over 1,800 objects represented by a large set of images from the THINGS database. GPT ratings were also compared with representations derived from a popular high performing deep convolutional network (AlexNet) trained on either the ImageNet database (referred to as AlexNet-LSVRC2012) or a more ecologically representative dataset (referred to as AlexNet-Ecoset; both from Mehrer et al., 2021, see Supplemental Figure 1).

We found that GPT-4o predicted SPOSE model distances (i.e., a model of human visual image ratings) reasonably well ($r_s$ = 0.545 based on the Carlson-Image stimuli; $r_s$ = 0.606 based on the THINGS stimuli; all Bonferroni-corrected $p < 0.05$; Supplemental Table 3), but overall the image-processing models aligned slightly less with human behavior (GPT-4-Vision: $r_s$ = 0.518 based on the Carlson-Image stimuli; $r_s$ = 0.529 based on the THINGS stimuli; GPT-4o-mini: $r_s$ = 0.533 based on the Carlson-Image stimuli; $r_s$ = 0.537 based on the THINGS stimuli) than their text-only counter parts (W = 75, $p < 0.039$, two-sided Wilcoxon Rank-Sum test between GPT-text model alignment with human data and GPT-vision model alignment and SPOSE model distances). For each model, alignment was slightly better for the THINGS stimuli, perhaps due to the more natural depictions of each image in that dataset (e.g., including natural backgrounds), which could have been a better match to the training data of the visual GPT-4 and GPT-4o models. The AlexNet-Ecoset models were less well aligned than the GPT-4o model ratings, but otherwise aligned reasonably well with the model of human behavior based on THINGS stimuli ($r_s$ = 0.564) than the Carlson-Image stimuli ($r_s$ = 0.407), and overall better predicted SPOSE distances than AlexNet-LSVRC2012 (based on THINGS: $r_s$ = 0.428; based on Carlson-Image: $r_s$ = 0.372).

Representational similarity analysis facilitates comparisons across input domains (like vision and language), which are useful for assessing the potential for modality-agnostic conceptual representations (which is considered to be a central feature of semantic knowledge; see Patterson et al., 2007 and Simanova et al., 2014 for discussion). For example, human ratings of these objects via *text* were well aligned with the SPOSE model distances, which were based on images ($r_s$ = 0.729).

GPT-4o (and the other visual processing GPT-4 LLMs) were less well aligned across text and image domains (for example, GPT-4o text-ratings correlated with Carlson-Image ratings: $r_s$ = 0.478 and with THINGS images: $r_s$ = 0.527; see Supplemental Table 3). Notably, GPT-4o ratings of these object *words* aligned better ($r_s$ = 0.686) with SPOSE ratings (which were derived based on images) than the GPT-4o model's ratings of images, as did GloVe ($r_s$ = 0.678), and Gemma2-9b ($r_s$ = 0.612). In some cases, model alignment with human pairwise *text* ratings increased given the reduced 55-item stimulus set that accommodated the THINGS dataset classes, but where possible we defer to the results of the larger 67-item sample.

Similar to the text experiments, Human-LLM alignment for images was explored with respect to within and between category ratings. Again, models were decidedly more aligned with human ratings for within-category comparisons (within-category GPT-4-Vision: based on Carlson-Image: $r_s$ = 0.564; based on THINGS: $r_s$ = 0.589; within-category GPT-4o-mini: based on Carlson-Image: $r_s$ = 0.458; based on THINGS: $r_s$ = 0.489; within-category GPT-4o: based on Carlson-Image: $r_s$ = 0.645; based on THINGS: $r_s$ = 0.592, all Bonferroni-corrected $p < 0.05$) than for between-category comparisons (GPT-4-Vision for Carlson-Image: $r_s$ = 0.250; GPT-4-Vision for THINGS: $r_s$ = 0.220; GPT-4o-mini for Carlson-Image: $r_s$ = 0.225; GPT-4o-mini for THINGS: $r_s$ = 0.196; GPT-4o for Carlson-Image: $r_s$ = 0.309; GPT-4o for THINGS: $r_s$ = 0.358, all Bonferroni-corrected $p < 0.05$; W = 144, $p < 0.001$, two-sided Wilcoxon Rank-Sum test for within vs between category GPT vision model alignment). AlexNet models were less aligned to SPOSE (human behavior) for both within-category (AlexNet-Ecoset for Carlson-Image: $r_s$ = 0.109; AlexNet-Ecoset for THINGS: $r_s$ = 0.469; AlexNet-LSVRC2012 for Carlson-Image: $r_s$ = 0.080; AlexNet-LSVRC2012 for THINGS: $r_s$ = 0.316, with only the THINGS analyses surviving multiple comparison correction) and between-category structure (AlexNet-Ecoset for Carlson-Image: $r_s$ = 0.051; AlexNet-Ecoset for THINGS: $r_s$ = 0.231; AlexNet-LSVRC2012 for Carlson-Image: $r_s$ = 0.060; AlexNet-LSVRC2012 for THINGS: $r_s$ = 0.024, with only the comparison with the AlexNet-Ecoset representations on the THINGS dataset surviving multiple comparison correction).

## Increasing human-LLM alignment through prompting and hyperparameters

LLMs provide increasingly accurate proxies for human ratings at the group level, especially for text-based tasks. However, there is interest in methods to increase the similarity between the representational and behavioral spaces of LLMs and humans (see Sucholutsky, et al. 2023), and there is room for improvement for even the most well-aligned LLMs observed in our study (see Supplemental Tables 2 and 3). Thus, we undertook an additional set of experiments that explored ways to increase alignment between LLM and human behavior via changes to hyperparameters and model prompts. GPT models were the focus for these experiments because of their popularity and overall good alignment performance.

First, we investigated increasing the alignment of image processing among the GPT-4 family of models. These models achieved modest alignment with human behavior, and the text-only ratings of these object concepts were more similar to models of human visual semantic behavior. Therefore, relying more heavily on the text processing capabilities of GPT-4o may increase alignment with human behavior. To test this, a new set of GPT-4 participants was run for the image rating task, where each LLM participant first provided a description of each image. Then, the LLM participants were asked to make their pairwise ratings based on their own *descriptions* of the images they had just provided (see Figure 1B and Example 3). In other words, the images were first converted to text descriptions, and similarity judgments were made based on these text descriptions. The results of these experiments are also indicated in Supplemental Tables 2 and 3 (denoted “GPT-4-Vision Descriptions” or “Vis. Desc.”). Rating text descriptions in this way increased alignment between each GPT-4 image processing model and the SPOSE model of human visual semantics for both stimulus sets (for GPT-4-Vision based on Carlson-Image: $r_s$ = 0.518 to 0.635; based on THINGS: $r_s$ = 0.529 to 0.613; for GPT-4o-mini based on Carlson-Image: $r_s$ = 0.533 to 0.592; based on THINGS: $r_s$ = 0.537 to 0.665; for GPT-4o based on Carlson-Image: $r_s$ = 0.545 to 0.610; based on THINGS: $r_s$ = 0.606 to 0.653; W = 1, $p < 0.005$, two-sided Wilcoxon Rank-Sum test comparing LLM image ratings and SPOSE model alignment with LLM text description ratings and SPOSE model alignment). Ratings based on text-descriptions of the images also increased cross-modal alignment for each model’s corresponding text-only word ratings, and this was especially notable for GPT-4o-mini (for GPT-4-Vision based on Carlson-Image: $r_s$ = 0.458 to 0.552; based on THINGS: $r_s$ = 0.504 to 0.559 for GPT-4o-mini based on Carlson-Image: $r_s$ = 0.490 to 0.586; based on THINGS: $r_s$ = 0.483 to 0.619; for GPT-4o based on Carlson-Image: $r_s$ = 0.478 to 0.526; based on THINGS: $r_s$ = 0.527 to 0.558; W = 1, $p < 0.005$, two-sided Wilcoxon Rank-Sum test).

Next, we assessed whether different approaches to operationalizing LLM participants impacted alignment results. Aher et al. (2023) ascribed a surname and honorific to each new LLM participant, and other experiments have simply re-queried the model without explicitly assigning an identity, and still obtained a distribution of human-like responses (Cava & Tagarelli, 2024; Marjieh et al., 2024; Mei et al., 2024) or did not find substantial differences among individual-level prompts for improving alignment with human behavior (Petrov et al., 2024). To examine what effect this has on LLM responses and the semantic representation distances elicited by our RSA experiment, we re-ran our experiments for the GPT models with all surnames and honorifics removed and measured group-level representational alignment. The results of these experiments are reported in Supplemental Tables 2 and 3 (denoted by “(Repeats)” or “(Rep.)”). Removing surnames and honorifics increased alignment with human ratings for each of the text-only GPT models (increase in GPT-3.5 alignment from $r_s$ = 0.456 to $r_s$ = 0.521; increase in GPT-4 alignment from $r_s$ = 0.696 to $r_s$ = 0.708; increase in GPT-4o-mini alignment from $r_s$

= 0.736 to $r_s$ = 0.746). GPT-4o run without surnames or honorifics produced the highest correlation between LLM and human responses observed in these studies (increase in GPT-4o alignment from $r_s$ = 0.740 to $r_s$ = 0.758). However, there were mixed results regarding whether this more minimal style of prompting improved alignment in the image rating experiments (see Supplemental Table 3).

The temperature hyperparameter increases or decreases the verbosity and randomness of a large language model's responses, and thus could influence LLM participant responses (as reported by Cava and Tagarelli, 2024). We explored the influence of this hyperparameter in our text-based experiments by re-running our GPT-3.5 and GPT-4 participants across a range of temperature settings: 0.01, 0.7 and 1.5 (combined with our results reported thus far which were run at 1.0). In general, this did not have a substantial or systematic influence on alignment (GPT-3.5 $r_s$ = 0.451, 0.464, 0.456, 0.454; and GPT-4 $r_s$ = 0.687, 0.692, 0.696, 0.699 across temperatures of 0.01, 0.7, 1.0 and 1.5, respectively). However, we noted changes in the consistency of responses across participants in these temperature-sweep experiments, which we address in the following section.

## Individual variability in human and LLM responses

Individual differences (or the variability observed between individuals) is a fundamental aspect of human behavior. Inter-individual variability was salient in our human behavioral data (for text-based ratings, Figure 2A), and overall lead to only modest agreement among human raters. Thus, an accurate encapsulation of human behavior by a cohort of LLM participants would be able to achieve *both* a high alignment with human responses at the group level, and moderate inter-individual variability. Quantifying and matching these inter-individual differences with LLMs is critical for generating useful proxies of human behavior, but this has thus far been a less-explored dimension of human-AI alignment. For example, when viewing a cohort of GPT-4o participants' responses (Figure 2B) there is a stark contrast in terms of the homogeneity of LLM responses relative to human participants. To quantify this aspect of performance, alignment among the unique participants (human and simulated) within each experiment was calculated along with an intraclass correlation coefficient (ICC) among each set of participants. An additional analysis examined how well a given LLM participant might have aligned with an individual human participant (i.e., individual-level human-LLM alignment). We compare these measures alongside group-level representational alignment in Figure 2.

LLMs that achieved the best overall alignment with human data at the group level produced strikingly consistent response patterns and were much more consistent than the cohort of human raters. The distribution of responses varied widely across models (Supplemental Figure 2). Human data (both text ratings and SPOSE distances) had a slightly bimodal distribution of rating responses. This feature was best captured by the GPT-4o and Gemma models but models generally lacked this characteristic. Changes in hyperparameters or prompting, particularly temperature, influenced inter-subject agreement via both ICC and inter-subject alignment (Figure 2C). However, there was only a small influence on group level alignment with human data. Overall GPT-4, GPT-4o-mini and GPT-4o were much more consistent in responses than human participants. GPT-4o achieved strong group-level alignment with human judgments ($r_s$ = 0.740 to 0.758), which exceeded the range of individual human inter-subject correlations ($r_s$ among pairs of individual human subjects ranged from 0.363 to 0.654, median 0.504). GPT-3.5 (*ICC* = 0.58 to 0.71), GPT-4o ratings of images (*ICC* = 0.44 to 0.50) and Gemma-7b (*ICC* = 0.42) models were closer to capturing the inter-subject agreement of human participants (*ICC* = 0.49), but overall, their alignment with human behavior (at the group level) was lower ($r_s$ = 0.502 to 0.606).

Finally, individual-level alignment between human and LLM participants was evaluated to understand how closely a given LLM participant might align with a given human participant's behavioral ratings. Figure 2D displays the alignment between each LLM participant (organized by model system) and each human participant (colored points). The highest individual human-LLM alignment was observed between a GPT-4o-mini participant and human participant S5 ($r_s$ = 0.633). However, overall, inter-individual cross-system alignment was lower than when responses from LLM and human participants were averaged ($r_s$ = 0.633 best individual alignment compared to $r_s$ = 0.746 for group-averaged GPT-4o-mini alignment). However, while no model seems to perfectly represent a *specific* human's performance, the performance of a given model may be overall similar to *another* human: many LLM participants (GPT-4 and above, Gemma2, Solar) aligned with individual human participants in a range that is similar to the alignment observed among individual human participants ($r_s$ = 0.363 to 0.654). Participants from each model also appeared to occupy distinct spaces from one another when rating distances were visualized in a common low dimensional space (Supplemental Figures 3 and 4). In general, no models overlapped with the human participant responses, which were mostly clustered tightly together (although Phi-3-medium and GPT-4o-mini participants were nearby).

## Discussion

The performance of LLMs continues to rapidly improve, raising increasingly important questions about reliability, explainability and alignment with human objectives. If LLMs begin to be used widely as proxies for human behavior (potentially in simulations, as assistants or for human subjects testing), methods will be needed for assessing how human-like a given model's behavior can be across a wide array of scenarios and for increasing alignment between LLMs and humans. We adapted a generalizable pairwise rating task, based around RSA, to probe the representational structure of LLMs that would otherwise be black-box interfaces. Experiments using this task found that GPT-4o's ratings conveyed a representational structure that is highly (but not perfectly) aligned with human semantic representational structure (obtained through the same behavioral pairwise rating task), especially when compared to other smaller models and when relying primarily on text processing capabilities (regardless of the input modality). Also, despite being smaller, many of the recent generation of compact 8- to 14-billion parameter language models such as Llama-3, Phi-3, Gemma-2 and Solar were still well aligned with human semantic ratings (more even than the substantially larger GPT-3.5). However, the inter-individual variability observed among humans was difficult to reproduce among LLM participants. Group-level alignment between LLMs and human behavior could be increased by changing some prompts and hyperparameters. Finally, human participants' object ratings were more consistent across text and image modalities than cross-modal LLM ratings.

These studies extend prior work (Dickson et al., 2024; Marjieh et al., 2022, 2024; Marjieh, Rijn, et al., 2023) by examining multiple models, comparisons within and across stimulus domains, and by examining inter-individual variability all within the same framework using matched stimuli. Our use of a continuous subjective rating task rather than assessing accuracy or performance like many LLM evaluations (e.g., Zhou et al., 2024) provides a useful complement to standard practices. The pairwise rating method used here can help probe nuanced, high-level features of knowledge representations and relationships among concepts in a flexible manner.

There are a number of limitations to this study that are well-suited for future work. First, our main goal was to develop a generalizable method for querying LLM behavior as a tool to understand Human-LLM alignment, reliability and explainability. These experiments involved a set of well-studied stimuli (words, images etc.) and relied heavily on previously collected or publicly available data sets. However, these materials, like all stimulus sets, are not exhaustive, are limited in scope and may have other shortcomings and biases (see Grootswagers and Robinson, 2021 or Thoret et al., 2021 for discussion). This of course stems in large part from their being developed with human testing in mind, and thus would be subject to practical constraints on a human participant's time and patience. A more comprehensive set of stimuli to fully probe LLM behavior and knowledge will require additional development. This might include scaling up behavioral RSA testing stimuli to operate over more sentences, paragraphs, audio clips or movies, as well as using stimuli that can better target expertise, emotions or personality dimensions.

A related limitation is that our human comparison cohort comprised a relatively small sample of participants ($n$ = 8). As discussed, pairwise rating tasks are burdensome to carry out due to the time and effort required of participants. Nonetheless, obtaining behavioral ratings for a larger number of stimuli from a larger number of human participants will be useful for grounding and expanding future explorations of LLM knowledge. Borrowing the large-scale testing structure of Hebart and colleagues (2020) could be useful in future endeavors for obtaining these data. Indeed, the work by Du and toher (2025) shows this can be effective approach. Pairwise ratings are a powerful tool in psychological research for probing knowledge and perceptual representations, but can be a prohibitively time and resource intensive approach. For example, they require all pairs of items in the probe set to be rated (or at least all unique pairs of non-identical items), and the number of trials required in these studies increases substantially with each new item added to the set (Dickson et al., 2024; Giordano et al., 2011). While there are some more efficient paradigms that can approximate pairwise ratings (e.g. Giordano et al., 2011; Glasgow et al., 2017; Hebart et al., 2020), LLMs could eventually be integrated into paradigm and stimulus development to test, norm or automatically generate near-human-quality similarity ratings without the time, money or effort required to elicit ratings from human participants. This could accelerate some aspects of psychological research and allow for a more rapid search of optimal paradigms, stimuli or psychologically useful feature spaces (as suggested by Dickson et al., 2024).

Future work might also explore behavioral RSA approaches with respect to different kinds of behavioral context or adapt it to more naturalistic interactions. Indeed, one potential limitation of these initial experiments is that they often compared representations of words and images presented in isolation (without context). This may have disadvantaged some models that do not support a chat-prompt interface and explicitly rely on the surrounding linguistic context for word representations (e.g., the BERT family of models). However, it is clear that humans can compare semantic relations among words in isolation (Carlson et al., 2014; Hansen & Hebart, 2022; Jiang et al., 2022), so this method of comparison has some obvious validity. Nonetheless, future work could benefit from querying ratings for stimuli within a more naturalistic (e.g., interactive or conversational) context. Similarly, an important next step will be assessing pairwise rating tasks relative to more diverse kinds of behavioral tasks or outputs (potentially beyond or in addition to the pairwise ratings studied here).

Follow up studies might use RSA to directly compare the representational structure of LLM *embeddings* across layers (during task performance, as a query is processed) and the model's *behavior* or task outputs. This would provide further insight into LLM knowledge representations and reasoning, and could illuminate processes related to LLM hallucinations (Tonmoy et al., 2024). A suitable analogy might be adapting a psychological experiment that measures behavior to a cognitive neuroscience study that measures neural processing during task performance (via EEG or fMRI). This approach could query how representational structure changes throughout the model's architecture and into deeper levels of processing (again similar to Cichy, Khosla, et al., 2016, and Carlson et al., 2014, studying representations across neural regions). This would require white box access to model activations or an ability to read out responses across layers for every query, which may be difficult for some frontier models where these are not made available.

Pairwise rating tasks, RSA and related techniques can be used in the service of increasing alignment between AIs and humans. One approach involves incorporating human-like representational knowledge as an objective function to improve alignment during training or fine-tuning (see Zhao et al., 2025 for very promising initial work on this approach), or fine-tuning models specifically to emulate human behavior in cognitive testing (e.g., Binz et al., 2025). Some modifications to our tasks and prompts were able to improve alignment between LLM responses and human data, but these strategies may not scale and were posed primarily as empirical questions to better characterize Turing Experimentation. More direct approaches aimed at improving alignment could be realized by incorporating objective functions that account for pairwise dissimilarity ratings provided by humans (see Sucholutsky et al., 2023 for summary and discussion). A particularly useful direction could be to develop models that can be directly aligned not only to the representational space of the general public, but also to that of experts in a particular area. Here, distinctions among specific within-versus-between category representations for a set of items could be useful analytical distinctions.

## Conclusion

We developed a generalizable Human-LLM alignment task based on pairwise ratings that allowed us to systematically compare 15 language models ranging from seven to at least hundreds of billions of parameters across text and image tasks. GPT-4o achieved the highest correlation with human semantic judgments (as high as $r_s$ = 0.758 for text, $r_s$ = 0.606 for images). Capturing the inter-individual variability of human behavior is still an outstanding issue among the LLMs evaluated here, and no model adequately captured this dimension of behavior while delivering high group-level alignment with human ratings.

The pairwise rating method adapted here for LLM evaluation comprises a quantitative framework for measuring alignment between human and artificial intelligence across input modalities. This method leverages an established cognitive science method and enabled us to describe some strengths and limitations of current LLMs: strong group-level semantic alignment but poor encapsulation of individual differences. This approach bridges cognitive science and AI evaluation, offering a scalable method that can assess how artificial systems encode knowledge and align with human cognition.

## Acknowledgments

We acknowledge support from the Independent Research and Development (IRAD) Fund from the Research and Exploratory Development Mission Area of the Johns Hopkins Applied Physics Laboratory. We thank L. Robert Slevc and his laboratory for collecting and sharing the human behavioral data and we thank Will Coon for helpful comments and discussions while drafting this manuscript. Feedback from Claude (hosted by Anthropic) was used to update some of the verbiage and language in this report.

## Significance Statement

The possibility of having artificial intelligence integrated into societal roles requires methods for understanding how these models align with a human understanding of the world. We developed a generalizable testing framework based on a classic cognitive psychology technique that elicits pairwise ratings for sets of stimuli and we used this method to measure alignment between human and artificial intelligence for text and image inputs. We found that modern frontier large language models aligned best with human ratings on this task but no model we tested accurately reproduced the individual variability observed in human cohorts. This approach opens the aperture for understanding and bridging the knowledge representations of human and artificial intelligence.

## Data Availability

Experimental materials, code (used for data collection and analysis) and data are available online: [repository link to come]

# Appendix

## Supplemental Table 1

*Supplemental Table 1* Word and image stimuli used In our experiments and the datasets they originated from.

| Object | Class | Original (Carlson-Image) | THINGS Image |
|---|---|---|---|
| hand | Human | stimulus1.png | hand_10s.jpg |
| ear | Human | stimulus2.png | ear_04s.jpg |
| chef | Human | - | - |
| hair | Human | stimulus5.png | hair_08s.jpg |
| dancer | Human | - | - |
| woman | Human | stimulus23.png | woman_06s.jpg |
| eye | Human | stimulus9.png | eye_13s.jpg |
| man | Human | stimulus24.png | man_03s.jpg |
| finger | Human | stimulus11.png | finger_03s.jpg |
| fist | Human | - | - |
| child | Human | stimulus21.png | baby_14s.jpg |
| armadillo | Animal | - | - |
| camel | Animal | stimulus26.png | camel_02s.jpg |
| snake | Animal | stimulus27.png | snake_04s.jpg |
| wolf | Animal | stimulus28.png | wolf_10s.jpg |
| monkey | Animal | stimulus29.png | monkey_13s.jpg |
| ostrich | Animal | stimulus30.png | ostrich_17s.jpg |
| zebra | Animal | stimulus32.png | zebra_03s.jpg |
| elephant | Animal | stimulus33.png | elephant_02n.jpg |
| sheep | Animal | stimulus35.png | sheep_02s.jpg |
| frog | Animal | stimulus36.png | frog_03s.jpg |
| cow | Animal | stimulus37.png | cow_08s.jpg |
| goat | Animal | stimulus38.png | goat_04s.jpg |
| dog | Animal | stimulus42.png | dog_02s.jpg |
| alligator | Animal | stimulus45.png | alligator_07s.jpg |
| giraffe | Animal | stimulus46.png | giraffe_03s.jpg |
| lion | Animal | stimulus47.png | lion_13s.jpg |
| carrot | Nautral Objects | stimulus49.png | carrot_01b.jpg |
| grape | Nautral Objects | stimulus50.png | grape_04s.jpg |
| potato | Nautral Objects | stimulus51.png | potato_13n.jpg |
| tree | Nautral Objects | stimulus67.png | tree_02s.jpg |
| pepper | Nautral Objects | stimulus70.png | pepper2_06s.jpg |
| lettuce | Nautral Objects | stimulus54.png | lettuce_04n.jpg |
| kiwi | Nautral Objects | stimulus55.png | kiwi_01b.jpg |
| cucumber | Nautral Objects | stimulus56.png | cucumber_04s.jpg |
| leaf | Nautral Objects | stimulus57.png | leaf_06s.jpg |
| apple | Nautral Objects | stimulus58.png | apple_01b.jpg |
| radish | Nautral Objects | stimulus59.png | radish_10s.jpg |
| eggplant | Nautral Objects | stimulus60.png | eggplant_07s.jpg |
| lake | Nautral Objects | - | - |
| pinecone | Nautral Objects | stimulus62.png | pinecone_10s.jpg |
| banana | Nautral Objects | stimulus63.png | banana_12s.jpg |
| tomato | Nautral Objects | stimulus64.png | tomato_12s.jpg |
| garlic | Nautral Objects | stimulus65.png | garlic_08n.jpg |
| path | Nautral Objects | - | - |
| pineapple | Nautral Objects | stimulus68.png | pineapple_07s.jpg |
| pear | Nautral Objects | stimulus69.png | pear_02s.jpg |
| waterfall | Nautral Objects | - | - |
| city | Manmade Objects | - | - |
| bottle | Manmade Objects | stimulus73.png | bottle_10s.jpg |
| lightbulb | Manmade Objects | stimulus74.png | lightbulb_03s.jpg |
| sign | Manmade Objects | - | - |
| cassette | Manmade Objects | stimulus76.png | cassette_05s.jpg |
| church | Manmade Objects | - | - |
| flag | Manmade Objects | stimulus78.png | flag_11s.jpg |
| key | Manmade Objects | stimulus79.png | key_01b.jpg |
| pliers | Manmade Objects | stimulus80.png | pliers_05s.jpg |
| arch | Manmade Objects | stimulus81.png | arch_04s.jpg |
| door | Manmade Objects | stimulus82.png | door_11s.jpg |
| hammer | Manmade Objects | stimulus83.png | hammer_07s.jpg |
| chair | Manmade Objects | stimulus84.png | chair_04s.jpg |
| gun | Manmade Objects | stimulus85.png | gun_05s.jpg |
| house | Manmade Objects | - | - |
| dome | Manmade Objects | - | - |
| umbrella | Manmade Objects | stimulus88.png | umbrella_09s.jpg |
| phone | Manmade Objects | stimulus89.png | phone_04s.jpg |
| stove | Manmade Objects | stimulus91.png | stove1_09s.jpg |

# Supplemental Table 2

*Supplemental Table 2* Representational alignment (measured by Spearman correlation between DSMs) across models for the set of 67 object words from Carlson et al., (2014), averaged at the group level where appropriate. Distance metrics were re-oriented to f

| | Human (Word Rating) | GPT-3.5 | GPT-3.5 (Repeats) | GPT-4 | GPT-4 (Repeats) | GPT-4o-mini | GPT-4o-mini (Repeats) | GPT-4o | GPT-4o (Repeats) | Gemma-7b | Gemma-2-9b | Phi-3-Medium-14b | Mistral-7b | Solar-10.7b | Llama2-7b | Llama2-uncensored-7b | Llama3-8b | bert-base-uncased | GloVe | albert-xxlarge-v2 | Ada Embeddings |
|---|---|---|---|---|---|---|---|---|---|---|---|---|---|---|---|---|---|---|---|---|---|
| **Human (Word Rating)** | 1.000 | 0.456 | 0.521 | 0.696 | 0.708 | 0.736 | 0.746 | 0.740 | 0.758 | 0.502 | 0.658 | 0.586 | 0.532 | 0.647 | 0.245 | 0.201 | 0.586 | 0.188 | 0.643 | - | 0.437 |
| **GPT-3.5** | 0.456 | 1.000 | 0.882 | 0.605 | 0.632 | 0.700 | 0.672 | 0.602 | 0.586 | 0.715 | 0.702 | 0.720 | 0.610 | 0.534 | 0.500 | 0.285 | 0.679 | 0.206 | 0.441 | 0.103 | 0.258 |
| **GPT-3.5 (Repeats)** | 0.521 | 0.882 | 1.000 | 0.660 | 0.681 | 0.753 | 0.725 | 0.656 | 0.644 | 0.729 | 0.741 | 0.748 | 0.656 | 0.590 | 0.494 | 0.279 | 0.674 | 0.221 | 0.508 | - | 0.298 |
| **GPT-4** | 0.696 | 0.605 | 0.660 | 1.000 | 0.965 | 0.827 | 0.833 | 0.865 | 0.862 | 0.579 | 0.834 | 0.686 | 0.609 | 0.709 | 0.367 | 0.228 | 0.694 | 0.181 | 0.669 | - | 0.402 |
| **GPT-4 (Repeats)** | 0.708 | 0.632 | 0.681 | 0.965 | 1.000 | 0.846 | 0.854 | 0.871 | 0.872 | 0.610 | 0.844 | 0.708 | 0.624 | 0.711 | 0.381 | 0.235 | 0.708 | 0.194 | 0.683 | - | 0.415 |
| **GPT-4o-mini** | 0.736 | 0.700 | 0.753 | 0.827 | 0.846 | 1.000 | 0.968 | 0.842 | 0.848 | 0.700 | 0.847 | 0.785 | 0.686 | 0.726 | 0.445 | 0.293 | 0.728 | 0.221 | 0.650 | - | 0.413 |
| **GPT-4o-mini (Repeats)** | 0.746 | 0.672 | 0.725 | 0.833 | 0.854 | 0.968 | 1.000 | 0.853 | 0.859 | 0.675 | 0.849 | 0.766 | 0.679 | 0.729 | 0.430 | 0.286 | 0.717 | 0.214 | 0.655 | - | 0.404 |
| **GPT-4o** | 0.740 | 0.602 | 0.656 | 0.865 | 0.871 | 0.842 | 0.853 | 1.000 | 0.941 | 0.592 | 0.825 | 0.683 | 0.624 | 0.712 | 0.371 | 0.250 | 0.686 | 0.240 | 0.700 | - | 0.420 |
| **GPT-4o (Repeats)** | 0.758 | 0.586 | 0.644 | 0.862 | 0.872 | 0.848 | 0.859 | 0.941 | 1.000 | 0.584 | 0.809 | 0.683 | 0.611 | 0.711 | 0.350 | 0.239 | 0.675 | 0.239 | 0.715 | - | 0.428 |
| **Gemma-7b** | 0.502 | 0.715 | 0.729 | 0.579 | 0.610 | 0.700 | 0.675 | 0.592 | 0.584 | 1.000 | 0.719 | 0.687 | 0.625 | 0.569 | 0.448 | 0.282 | 0.623 | 0.243 | 0.486 | - | 0.318 |
| **Gemma-2-9b** | 0.658 | 0.702 | 0.741 | 0.834 | 0.844 | 0.847 | 0.849 | 0.825 | 0.809 | 0.719 | 1.000 | 0.742 | 0.682 | 0.728 | 0.444 | 0.285 | 0.735 | 0.253 | 0.639 | - | 0.396 |
| **Phi-3-Medium-14b** | 0.586 | 0.720 | 0.748 | 0.686 | 0.708 | 0.785 | 0.766 | 0.683 | 0.683 | 0.687 | 0.742 | 1.000 | 0.657 | 0.597 | 0.471 | 0.297 | 0.692 | 0.172 | 0.465 | - | 0.257 |
| **Mistral-7b** | 0.532 | 0.610 | 0.656 | 0.609 | 0.624 | 0.686 | 0.679 | 0.624 | 0.611 | 0.625 | 0.682 | 0.657 | 1.000 | 0.620 | 0.403 | 0.310 | 0.602 | 0.134 | 0.450 | - | 0.261 |
| **Solar-10.7b** | 0.647 | 0.534 | 0.590 | 0.709 | 0.711 | 0.726 | 0.729 | 0.712 | 0.711 | 0.569 | 0.728 | 0.597 | 0.620 | 1.000 | 0.314 | 0.240 | 0.598 | 0.228 | 0.649 | - | 0.456 |
| **Llama2-7b** | 0.245 | 0.500 | 0.494 | 0.367 | 0.381 | 0.445 | 0.430 | 0.371 | 0.350 | 0.448 | 0.444 | 0.471 | 0.403 | 0.314 | 1.000 | 0.293 | 0.428 | - | 0.229 | - | 0.146 |
| **Llama2-uncensored-7b** | 0.201 | 0.285 | 0.279 | 0.228 | 0.235 | 0.293 | 0.286 | 0.250 | 0.239 | 0.282 | 0.285 | 0.297 | 0.310 | 0.240 | 0.293 | 1.000 | 0.298 | 0.102 | 0.137 | - | - |
| **Llama3-8b** | 0.586 | 0.679 | 0.674 | 0.694 | 0.708 | 0.728 | 0.717 | 0.686 | 0.675 | 0.623 | 0.735 | 0.692 | 0.602 | 0.598 | 0.428 | 0.298 | 1.000 | 0.095 | 0.527 | 0.093 | 0.359 |
| **bert-base-uncased** | 0.188 | 0.206 | 0.221 | 0.181 | 0.194 | 0.221 | 0.214 | 0.240 | 0.239 | 0.243 | 0.253 | 0.172 | 0.134 | 0.228 | - | 0.102 | 0.095 | 1.000 | 0.317 | - | 0.210 |
| **GloVe** | 0.643 | 0.441 | 0.508 | 0.669 | 0.683 | 0.650 | 0.655 | 0.700 | 0.715 | 0.486 | 0.639 | 0.465 | 0.450 | 0.649 | 0.229 | 0.137 | 0.527 | 0.317 | 1.000 | - | 0.590 |
| **albert-xxlarge-v2** | - | 0.103 | - | - | - | - | - | - | - | - | - | - | - | - | - | - | 0.093 | - | - | 1.000 | - |
| **Ada Embeddings** | 0.437 | 0.258 | 0.298 | 0.402 | 0.415 | 0.413 | 0.404 | 0.420 | 0.428 | 0.318 | 0.396 | 0.257 | 0.261 | 0.456 | 0.146 | - | 0.359 | 0.210 | 0.590 | - | 1.000 |
| **GPT-4-Vision Carlson-Image** | 0.371 | 0.194 | 0.231 | 0.382 | 0.380 | 0.299 | 0.317 | 0.344 | 0.365 | 0.164 | 0.287 | 0.259 | 0.208 | 0.340 | 0.088 | - | 0.290 | - | 0.379 | - | 0.273 |
| **GPT-4-Vision Carlson-Image (Repeats)** | 0.396 | 0.205 | 0.253 | 0.394 | 0.390 | 0.329 | 0.341 | 0.360 | 0.384 | 0.166 | 0.296 | 0.264 | 0.242 | 0.360 | - | - | 0.292 | - | 0.405 | - | 0.299 |
| **GPT-4-Vis. Desc. Carlson-Image** | 0.411 | 0.260 | 0.293 | 0.474 | 0.480 | 0.383 | 0.394 | 0.435 | 0.461 | 0.223 | 0.373 | 0.337 | 0.269 | 0.392 | 0.127 | - | 0.366 | - | 0.454 | - | 0.319 |
| **GPT-4-Vis. Desc. Carlson-Image (Repeats)** | 0.424 | 0.266 | 0.305 | 0.472 | 0.482 | 0.381 | 0.394 | 0.429 | 0.452 | 0.229 | 0.376 | 0.338 | 0.274 | 0.400 | 0.139 | - | 0.359 | - | 0.473 | - | 0.321 |
| **GPT-4o-mini Carlson-Image** | 0.442 | 0.334 | 0.368 | 0.495 | 0.496 | 0.458 | 0.471 | 0.489 | 0.495 | 0.314 | 0.451 | 0.396 | 0.350 | 0.453 | 0.199 | 0.100 | 0.424 | - | 0.446 | - | 0.304 |
| **GPT-4o-mini Carlson-Image (Repeats)** | 0.436 | 0.336 | 0.364 | 0.534 | 0.531 | 0.471 | 0.485 | 0.501 | 0.510 | 0.337 | 0.478 | 0.397 | 0.338 | 0.441 | 0.212 | 0.125 | 0.440 | - | 0.426 | - | 0.283 |
| **GPT-4o-mini Desc. Carlson-Image** | 0.525 | 0.376 | 0.429 | 0.532 | 0.546 | 0.520 | 0.527 | 0.531 | 0.539 | 0.390 | 0.506 | 0.475 | 0.374 | 0.446 | 0.263 | 0.130 | 0.465 | 0.129 | 0.507 | - | 0.344 |
| **GPT-4o-mini Desc. Carlson-Image (Repeats)** | 0.526 | 0.379 | 0.428 | 0.519 | 0.537 | 0.520 | 0.528 | 0.514 | 0.529 | 0.383 | 0.499 | 0.465 | 0.363 | 0.462 | 0.268 | 0.123 | 0.463 | 0.106 | 0.521 | - | 0.364 |
| **GPT-4o Carlson-Image** | 0.396 | 0.289 | 0.325 | 0.401 | 0.414 | 0.373 | 0.379 | 0.374 | 0.404 | 0.275 | 0.337 | 0.318 | 0.246 | 0.358 | 0.111 | - | 0.358 | - | 0.401 | 0.084 | 0.300 |
| **GPT-4o Carlson-Image (Repeats)** | 0.378 | 0.321 | 0.340 | 0.412 | 0.418 | 0.378 | 0.386 | 0.396 | 0.402 | 0.317 | 0.382 | 0.349 | 0.267 | 0.388 | 0.163 | - | 0.383 | - | 0.391 | - | 0.280 |
| **GPT-4o Desc. Carlson-Image** | 0.430 | 0.258 | 0.312 | 0.463 | 0.463 | 0.365 | 0.370 | 0.421 | 0.442 | 0.245 | 0.364 | 0.328 | 0.303 | 0.404 | 0.117 | - | 0.350 | - | 0.490 | - | 0.302 |
| **GPT-4o Desc. Carlson-Image (Repeats)** | 0.455 | 0.228 | 0.296 | 0.427 | 0.434 | 0.351 | 0.362 | 0.405 | 0.419 | 0.222 | 0.339 | 0.299 | 0.288 | 0.381 | 0.112 | - | 0.331 | - | 0.496 | - | 0.324 |

| | GPT-4-Vision Carlson-Image | GPT-4-Vision Carlson-Image (Repeats) | GPT-4-Vis. Desc. Carlson-Image | GPT-4-Vis. Desc. Carlson-Image (Repeats) | GPT-4o-mini Carlson-Image | GPT-4o-mini Carlson-Image (Repeats) | GPT-4o-mini Desc. Carlson-Image | GPT-4o-mini Desc. Carlson-Image (Repeats) | GPT-4o Carlson-Image | GPT-4o Carlson-Image (Repeats) | GPT-4o Desc. Carlson-Image | GPT-4o Desc. Carlson-Image (Repeats) |
|---|---|---|---|---|---|---|---|---|---|---|---|---|
| **Human (Word Rating)** | 0.371 | 0.396 | 0.411 | 0.424 | 0.442 | 0.436 | 0.525 | 0.526 | 0.396 | 0.378 | 0.430 | 0.455 |
| **GPT-3.5** | 0.194 | 0.205 | 0.260 | 0.266 | 0.334 | 0.336 | 0.376 | 0.379 | 0.289 | 0.321 | 0.258 | 0.228 |
| **GPT-3.5 (Repeats)** | 0.231 | 0.253 | 0.293 | 0.305 | 0.368 | 0.364 | 0.429 | 0.428 | 0.325 | 0.340 | 0.312 | 0.296 |
| **GPT-4** | 0.382 | 0.394 | 0.474 | 0.472 | 0.495 | 0.534 | 0.532 | 0.519 | 0.401 | 0.412 | 0.463 | 0.427 |
| **GPT-4 (Repeats)** | 0.380 | 0.390 | 0.480 | 0.482 | 0.496 | 0.531 | 0.546 | 0.537 | 0.414 | 0.418 | 0.463 | 0.434 |
| **GPT-4o-mini** | 0.299 | 0.329 | 0.383 | 0.381 | 0.458 | 0.471 | 0.520 | 0.520 | 0.373 | 0.378 | 0.365 | 0.351 |
| **GPT-4o-mini (Repeats)** | 0.317 | 0.341 | 0.394 | 0.394 | 0.471 | 0.485 | 0.527 | 0.528 | 0.379 | 0.386 | 0.370 | 0.362 |
| **GPT-4o** | 0.344 | 0.360 | 0.435 | 0.429 | 0.489 | 0.501 | 0.531 | 0.514 | 0.374 | 0.396 | 0.421 | 0.405 |
| **GPT-4o (Repeats)** | 0.365 | 0.384 | 0.461 | 0.452 | 0.495 | 0.510 | 0.539 | 0.529 | 0.404 | 0.402 | 0.442 | 0.419 |
| **Gemma-7b** | 0.164 | 0.166 | 0.223 | 0.229 | 0.314 | 0.337 | 0.390 | 0.383 | 0.275 | 0.317 | 0.245 | 0.222 |
| **Gemma-2-9b** | 0.287 | 0.296 | 0.373 | 0.376 | 0.451 | 0.478 | 0.506 | 0.499 | 0.337 | 0.382 | 0.364 | 0.339 |
| **Phi-3-Medium-14b** | 0.259 | 0.264 | 0.337 | 0.338 | 0.396 | 0.397 | 0.475 | 0.465 | 0.318 | 0.349 | 0.328 | 0.299 |
| **Mistral-7b** | 0.208 | 0.242 | 0.269 | 0.274 | 0.350 | 0.338 | 0.374 | 0.363 | 0.246 | 0.267 | 0.303 | 0.288 |
| **Solar-10.7b** | 0.340 | 0.360 | 0.392 | 0.400 | 0.453 | 0.441 | 0.446 | 0.462 | 0.358 | 0.388 | 0.404 | 0.381 |
| **Llama2-7b** | 0.088 | - | 0.127 | 0.139 | 0.199 | 0.212 | 0.263 | 0.268 | 0.111 | 0.163 | 0.117 | 0.112 |
| **Llama2-uncensored-7b** | - | - | - | - | 0.100 | 0.125 | 0.130 | 0.123 | - | - | - | - |
| **Llama3-8b** | 0.290 | 0.292 | 0.366 | 0.359 | 0.424 | 0.440 | 0.465 | 0.463 | 0.358 | 0.383 | 0.350 | 0.331 |
| **bert-base-uncased** | - | - | - | - | - | - | 0.129 | 0.106 | - | - | - | - |
| **GloVe** | 0.379 | 0.405 | 0.454 | 0.473 | 0.446 | 0.426 | 0.507 | 0.521 | 0.401 | 0.391 | 0.490 | 0.496 |
| **albert-xxlarge-v2** | - | - | - | - | - | - | - | - | 0.084 | - | - | - |
| **Ada Embeddings** | 0.273 | 0.299 | 0.319 | 0.321 | 0.304 | 0.283 | 0.344 | 0.364 | 0.300 | 0.280 | 0.302 | 0.324 |
| **GPT-4-Vision Carlson-Image** | 1.000 | 0.628 | 0.617 | 0.618 | 0.534 | 0.547 | 0.492 | 0.509 | 0.559 | 0.517 | 0.594 | 0.551 |
| **GPT-4-Vision Carlson-Image (Repeats)** | 0.628 | 1.000 | 0.557 | 0.569 | 0.528 | 0.516 | 0.483 | 0.498 | 0.548 | 0.495 | 0.562 | 0.554 |
| **GPT-4-Vis. Desc. Carlson-Image** | 0.617 | 0.557 | 1.000 | 0.778 | 0.562 | 0.576 | 0.567 | 0.584 | 0.617 | 0.579 | 0.670 | 0.634 |
| **GPT-4-Vis. Desc. Carlson-Image (Repeats)** | 0.618 | 0.569 | 0.778 | 1.000 | 0.571 | 0.593 | 0.605 | 0.629 | 0.637 | 0.597 | 0.691 | 0.647 |
| **GPT-4o-mini Carlson-Image** | 0.534 | 0.528 | 0.562 | 0.571 | 1.000 | 0.798 | 0.575 | 0.582 | 0.562 | 0.553 | 0.542 | 0.501 |
| **GPT-4o-mini Carlson-Image (Repeats)** | 0.547 | 0.516 | 0.576 | 0.593 | 0.798 | 1.000 | 0.618 | 0.609 | 0.578 | 0.588 | 0.546 | 0.510 |
| **GPT-4o-mini Desc. Carlson-Image** | 0.492 | 0.483 | 0.567 | 0.605 | 0.575 | 0.618 | 1.000 | 0.910 | 0.618 | 0.584 | 0.624 | 0.655 |
| **GPT-4o-mini Desc. Carlson-Image (Repeats)** | 0.509 | 0.498 | 0.584 | 0.629 | 0.582 | 0.609 | 0.910 | 1.000 | 0.648 | 0.590 | 0.636 | 0.663 |
| **GPT-4o Carlson-Image** | 0.559 | 0.548 | 0.617 | 0.637 | 0.562 | 0.578 | 0.618 | 0.648 | 1.000 | 0.717 | 0.616 | 0.630 |
| **GPT-4o Carlson-Image (Repeats)** | 0.517 | 0.495 | 0.579 | 0.597 | 0.553 | 0.588 | 0.584 | 0.590 | 0.717 | 1.000 | 0.563 | 0.553 |
| **GPT-4o Desc. Carlson-Image** | 0.594 | 0.562 | 0.670 | 0.691 | 0.542 | 0.546 | 0.624 | 0.636 | 0.616 | 0.563 | 1.000 | 0.788 |
| **GPT-4o Desc. Carlson-Image (Repeats)** | 0.551 | 0.554 | 0.634 | 0.647 | 0.501 | 0.510 | 0.655 | 0.663 | 0.630 | 0.553 | 0.788 | 1.000 |

# Supplemental Table 3

*Supplemental Table 3* Representational alignment (measured by Spearman correlation between DSMs) across models for the set of 55 object words/images that overlap between from Carlson et al., (2014) and THINGS, averaged at the group level where appropriate.

| | Human (SPOSE) | Human (Word Rating) | GPT-4-Vision Carlson-Image | GPT-4-Vision Carlson-Image (Repeats) | GPT-4-Vis. Desc. Carlson-Image | GPT-4-Vis. Desc. Carlson-Image (Repeats) | GPT-4o-mini Carlson-Image | GPT-4o-mini Carlson-Image (Repeats) | GPT-4o-mini Desc. Carlson-Image | GPT-4o-mini Desc. Carlson-Image (Repeats) | GPT-4o Carlson-Image | GPT-4o Carlson-Image (Repeats) | GPT-4o Desc. Carlson-Image | GPT-4o Desc. Carlson-Image (Repeats) | GPT-4-Vision THINGS | GPT-4-Vision THINGS (Repeats) | GPT-4-Vis. Desc. THINGS | GPT-4-Vis. Desc. THINGS (Repeats) | GPT-4o-mini THINGS | GPT-4o-mini THINGS (Repeats) | GPT-4o-mini Desc. THINGS | GPT-4o-mini Desc. THINGS (Repeats) | GPT-4o THINGS | GPT-4o THINGS (Repeats) | GPT-4o Desc. THINGS | GPT-4o Desc. THINGS (Repeats) | Alex-Net-ecoset THINGS | Alex-Net-LSVRC2012 THINGS | Alex-Net-ecoset Carlson-Image | Alex-Net-LSVRC2012 Carlson-Image |
|---|---|---|---|---|---|---|---|---|---|---|---|---|---|---|---|---|---|---|---|---|---|---|---|---|---|---|---|---|---|---|
| **Human (SPOSE)** | 1.000 | 0.729 | 0.518 | 0.494 | 0.635 | 0.632 | 0.533 | 0.570 | 0.592 | 0.622 | 0.545 | 0.521 | 0.610 | 0.607 | 0.529 | 0.515 | 0.613 | 0.644 | 0.537 | 0.591 | 0.665 | 0.670 | 0.606 | 0.604 | 0.653 | 0.652 | 0.564 | 0.428 | 0.407 | 0.372 |
| **Human (Word Rating)** | 0.729 | 1.000 | 0.434 | 0.458 | 0.480 | 0.487 | 0.477 | 0.480 | 0.573 | 0.590 | 0.467 | 0.438 | 0.498 | 0.526 | 0.494 | 0.509 | 0.511 | 0.511 | 0.479 | 0.503 | 0.564 | 0.571 | 0.542 | 0.474 | 0.554 | 0.545 | 0.573 | 0.513 | 0.486 | 0.455 |
| **GPT-3.5** | 0.409 | 0.490 | 0.271 | 0.272 | 0.309 | 0.303 | 0.368 | 0.371 | 0.421 | 0.428 | 0.371 | 0.367 | 0.333 | 0.338 | 0.282 | 0.302 | 0.362 | 0.343 | 0.360 | 0.354 | 0.460 | 0.432 | 0.373 | 0.384 | 0.367 | 0.368 | 0.435 | 0.356 | 0.366 | 0.296 |
| **GPT-3.5 (Repeats)** | 0.439 | 0.556 | 0.303 | 0.319 | 0.336 | 0.334 | 0.389 | 0.395 | 0.487 | 0.489 | 0.412 | 0.388 | 0.385 | 0.407 | 0.305 | 0.338 | 0.382 | 0.371 | 0.368 | 0.369 | 0.480 | 0.456 | 0.421 | 0.403 | 0.401 | 0.396 | 0.451 | 0.396 | 0.399 | 0.330 |
| **GPT-4** | 0.682 | 0.725 | 0.458 | 0.468 | 0.552 | 0.547 | 0.535 | 0.575 | 0.591 | 0.588 | 0.499 | 0.484 | 0.556 | 0.551 | 0.504 | 0.494 | 0.559 | 0.549 | 0.513 | 0.569 | 0.608 | 0.601 | 0.571 | 0.582 | 0.581 | 0.584 | 0.547 | 0.486 | 0.434 | 0.393 |
| **GPT-4 (Repeats)** | 0.685 | 0.737 | 0.455 | 0.465 | 0.556 | 0.556 | 0.538 | 0.574 | 0.603 | 0.604 | 0.509 | 0.490 | 0.548 | 0.551 | 0.495 | 0.485 | 0.563 | 0.555 | 0.511 | 0.559 | 0.621 | 0.612 | 0.563 | 0.571 | 0.581 | 0.588 | 0.552 | 0.475 | 0.438 | 0.390 |
| **GPT-4o-mini** | 0.632 | 0.770 | 0.380 | 0.414 | 0.459 | 0.450 | 0.490 | 0.506 | 0.586 | 0.599 | 0.476 | 0.446 | 0.456 | 0.472 | 0.440 | 0.455 | 0.506 | 0.508 | 0.483 | 0.515 | 0.619 | 0.613 | 0.541 | 0.496 | 0.543 | 0.551 | 0.560 | 0.466 | 0.438 | 0.373 |
| **GPT-4o-mini (Repeats)** | 0.641 | 0.777 | 0.403 | 0.427 | 0.476 | 0.468 | 0.506 | 0.529 | 0.592 | 0.606 | 0.484 | 0.457 | 0.463 | 0.481 | 0.448 | 0.456 | 0.506 | 0.509 | 0.489 | 0.522 | 0.618 | 0.618 | 0.538 | 0.502 | 0.544 | 0.552 | 0.566 | 0.466 | 0.433 | 0.378 |
| **GPT-4o** | 0.686 | 0.766 | 0.430 | 0.441 | 0.522 | 0.523 | 0.528 | 0.545 | 0.592 | 0.585 | 0.478 | 0.473 | 0.526 | 0.535 | 0.474 | 0.485 | 0.506 | 0.504 | 0.499 | 0.540 | 0.576 | 0.575 | 0.527 | 0.528 | 0.558 | 0.558 | 0.514 | 0.428 | 0.408 | 0.368 |
| **GPT-4o (Repeats)** | 0.694 | 0.788 | 0.443 | 0.456 | 0.546 | 0.539 | 0.531 | 0.553 | 0.591 | 0.591 | 0.492 | 0.467 | 0.534 | 0.533 | 0.482 | 0.487 | 0.523 | 0.531 | 0.505 | 0.537 | 0.588 | 0.589 | 0.536 | 0.520 | 0.561 | 0.565 | 0.533 | 0.443 | 0.424 | 0.369 |
| **Gemma-7b** | 0.363 | 0.500 | 0.200 | 0.202 | 0.251 | 0.245 | 0.318 | 0.343 | 0.420 | 0.426 | 0.347 | 0.366 | 0.293 | 0.301 | 0.200 | 0.240 | 0.276 | 0.280 | 0.304 | 0.287 | 0.410 | 0.391 | 0.322 | 0.342 | 0.339 | 0.324 | 0.337 | 0.288 | 0.316 | 0.230 |
| **Gemma-2-9b** | 0.612 | 0.697 | 0.371 | 0.371 | 0.446 | 0.441 | 0.482 | 0.515 | 0.566 | 0.574 | 0.440 | 0.451 | 0.447 | 0.463 | 0.406 | 0.432 | 0.458 | 0.456 | 0.462 | 0.491 | 0.559 | 0.543 | 0.492 | 0.506 | 0.517 | 0.513 | 0.498 | 0.412 | 0.414 | 0.341 |
| **Phi-3-Medium-14b** | 0.488 | 0.620 | 0.332 | 0.335 | 0.382 | 0.377 | 0.426 | 0.436 | 0.534 | 0.524 | 0.401 | 0.392 | 0.395 | 0.398 | 0.359 | 0.361 | 0.425 | 0.405 | 0.428 | 0.441 | 0.506 | 0.501 | 0.444 | 0.417 | 0.402 | 0.426 | 0.456 | 0.451 | 0.411 | 0.378 |
| **Mistral-7b** | 0.395 | 0.548 | 0.255 | 0.298 | 0.299 | 0.286 | 0.349 | 0.338 | 0.411 | 0.405 | 0.305 | 0.300 | 0.360 | 0.369 | 0.277 | 0.323 | 0.300 | 0.287 | 0.317 | 0.306 | 0.365 | 0.351 | 0.356 | 0.324 | 0.330 | 0.331 | 0.370 | 0.324 | 0.303 | 0.265 |
| **Solar-10.7b** | 0.575 | 0.683 | 0.379 | 0.401 | 0.445 | 0.439 | 0.479 | 0.478 | 0.497 | 0.526 | 0.427 | 0.442 | 0.465 | 0.456 | 0.403 | 0.407 | 0.439 | 0.437 | 0.450 | 0.461 | 0.519 | 0.513 | 0.473 | 0.498 | 0.507 | 0.487 | 0.479 | 0.423 | 0.378 | 0.348 |
| **Llama2-7b** | 0.195 | 0.275 | 0.159 | 0.112 | 0.164 | 0.165 | 0.200 | 0.202 | 0.299 | 0.316 | 0.172 | 0.184 | 0.177 | 0.196 | 0.120 | 0.135 | 0.168 | 0.161 | 0.222 | 0.219 | 0.278 | 0.265 | 0.219 | 0.263 | 0.214 | 0.243 | 0.237 | 0.231 | 0.185 | 0.165 |
| **Llama2-uncensored-7b** | 0.131 | 0.202 | - | - | - | - | - | 0.117 | 0.145 | 0.142 | - | - | - | - | - | - | - | - | - | - | 0.159 | 0.161 | 0.113 | 0.124 | - | 0.116 | 0.155 | 0.145 | 0.147 | - |
| **Llama3-8b** | 0.567 | 0.599 | 0.367 | 0.360 | 0.427 | 0.401 | 0.449 | 0.475 | 0.493 | 0.497 | 0.423 | 0.412 | 0.422 | 0.415 | 0.387 | 0.395 | 0.433 | 0.437 | 0.457 | 0.485 | 0.523 | 0.504 | 0.460 | 0.504 | 0.470 | 0.481 | 0.496 | 0.439 | 0.397 | 0.367 |
| **bert-base-uncased** | - | 0.178 | - | - | - | - | - | - | 0.112 | - | - | - | - | - | - | - | - | - | - | - | - | - | - | - | - | - | - | - | - | - |
| **GloVe** | 0.678 | 0.699 | 0.427 | 0.441 | 0.521 | 0.523 | 0.479 | 0.480 | 0.549 | 0.564 | 0.444 | 0.450 | 0.532 | 0.535 | 0.439 | 0.466 | 0.466 | 0.488 | 0.466 | 0.471 | 0.549 | 0.550 | 0.470 | 0.477 | 0.553 | 0.532 | 0.466 | 0.314 | 0.335 | 0.257 |
| **albert-xxlarge-v2** | - | - | - | - | - | - | - | - | - | - | - | - | - | - | - | - | - | - | - | - | 0.122 | 0.128 | - | - | - | - | - | - | - | - |
| **Ada Embeddings** | 0.371 | 0.453 | 0.276 | 0.314 | 0.337 | 0.335 | 0.316 | 0.305 | 0.355 | 0.382 | 0.309 | 0.308 | 0.290 | 0.299 | 0.288 | 0.305 | 0.271 | 0.308 | 0.320 | 0.317 | 0.376 | 0.360 | 0.286 | 0.305 | 0.390 | 0.336 | 0.293 | 0.230 | 0.219 | 0.217 |
| **GPT-4-Vision Carlson-Image** | 0.518 | 0.434 | 1.000 | 0.670 | 0.661 | 0.665 | 0.595 | 0.612 | 0.534 | 0.551 | 0.607 | 0.560 | 0.628 | 0.597 | 0.511 | 0.478 | 0.516 | 0.543 | 0.529 | 0.551 | 0.470 | 0.456 | 0.522 | 0.503 | 0.478 | 0.500 | 0.437 | 0.423 | 0.411 | 0.430 |
| **GPT-4-Vision Carlson-Image (Repeats)** | 0.494 | 0.458 | 0.670 | 1.000 | 0.589 | 0.607 | 0.566 | 0.561 | 0.516 | 0.537 | 0.589 | 0.525 | 0.589 | 0.586 | 0.498 | 0.479 | 0.508 | 0.526 | 0.483 | 0.516 | 0.469 | 0.475 | 0.529 | 0.476 | 0.490 | 0.505 | 0.409 | 0.400 | 0.389 | 0.420 |
| **GPT-4-Vis. Desc. Carlson-Image** | 0.635 | 0.480 | 0.661 | 0.589 | 1.000 | 0.812 | 0.604 | 0.629 | 0.591 | 0.605 | 0.656 | 0.616 | 0.706 | 0.672 | 0.487 | 0.500 | 0.626 | 0.638 | 0.538 | 0.590 | 0.544 | 0.548 | 0.571 | 0.590 | 0.572 | 0.590 | 0.500 | 0.435 | 0.383 | 0.392 |
| **GPT-4-Vis. Desc. Carlson-Image (Repeats)** | 0.632 | 0.487 | 0.665 | 0.607 | 0.812 | 1.000 | 0.607 | 0.631 | 0.607 | 0.631 | 0.666 | 0.620 | 0.715 | 0.672 | 0.520 | 0.514 | 0.636 | 0.657 | 0.540 | 0.581 | 0.533 | 0.540 | 0.587 | 0.594 | 0.571 | 0.584 | 0.488 | 0.416 | 0.380 | 0.374 |
| **GPT-4o-mini Carlson-Image** | 0.533 | 0.477 | 0.595 | 0.566 | 0.604 | 0.607 | 1.000 | 0.831 | 0.588 | 0.604 | 0.609 | 0.592 | 0.579 | 0.553 | 0.497 | 0.458 | 0.526 | 0.532 | 0.697 | 0.677 | 0.562 | 0.561 | 0.530 | 0.512 | 0.498 | 0.505 | 0.473 | 0.454 | 0.432 | 0.440 |
| **GPT-4o-mini Carlson-Image (Repeats)** | 0.570 | 0.480 | 0.612 | 0.561 | 0.629 | 0.631 | 0.831 | 1.000 | 0.616 | 0.621 | 0.639 | 0.614 | 0.596 | 0.571 | 0.498 | 0.459 | 0.547 | 0.554 | 0.650 | 0.676 | 0.582 | 0.585 | 0.564 | 0.555 | 0.530 | 0.538 | 0.521 | 0.511 | 0.467 | 0.486 |
| **GPT-4o-mini Desc. Carlson-Image** | 0.592 | 0.573 | 0.534 | 0.516 | 0.591 | 0.607 | 0.588 | 0.616 | 1.000 | 0.912 | 0.628 | 0.577 | 0.647 | 0.684 | 0.478 | 0.491 | 0.581 | 0.566 | 0.537 | 0.585 | 0.674 | 0.672 | 0.598 | 0.561 | 0.568 | 0.607 | 0.495 | 0.455 | 0.464 | 0.466 |
| **GPT-4o-mini Desc. Carlson-Image (Repeats)** | 0.622 | 0.590 | 0.551 | 0.537 | 0.605 | 0.631 | 0.604 | 0.621 | 0.912 | 1.000 | 0.654 | 0.580 | 0.655 | 0.689 | 0.487 | 0.494 | 0.606 | 0.606 | 0.554 | 0.598 | 0.718 | 0.715 | 0.618 | 0.595 | 0.605 | 0.640 | 0.501 | 0.450 | 0.464 | 0.463 |
| **GPT-4o Carlson-Image** | 0.545 | 0.467 | 0.607 | 0.589 | 0.656 | 0.666 | 0.609 | 0.639 | 0.628 | 0.654 | 1.000 | 0.740 | 0.639 | 0.656 | 0.467 | 0.486 | 0.570 | 0.604 | 0.525 | 0.562 | 0.570 | 0.575 | 0.628 | 0.604 | 0.560 | 0.583 | 0.486 | 0.470 | 0.470 | 0.494 |
| **GPT-4o Carlson-Image (Repeats)** | 0.521 | 0.438 | 0.560 | 0.525 | 0.616 | 0.620 | 0.592 | 0.614 | 0.577 | 0.580 | 0.740 | 1.000 | 0.590 | 0.575 | 0.413 | 0.419 | 0.516 | 0.546 | 0.502 | 0.535 | 0.503 | 0.507 | 0.555 | 0.584 | 0.519 | 0.524 | 0.478 | 0.464 | 0.449 | 0.470 |
| **GPT-4o Desc. Carlson-Image** | 0.610 | 0.498 | 0.628 | 0.589 | 0.706 | 0.715 | 0.579 | 0.596 | 0.647 | 0.655 | 0.639 | 0.590 | 1.000 | 0.810 | 0.527 | 0.517 | 0.584 | 0.588 | 0.501 | 0.548 | 0.517 | 0.521 | 0.576 | 0.570 | 0.553 | 0.560 | 0.464 | 0.404 | 0.400 | 0.390 |
| **GPT-4o Desc. Carlson-Image (Repeats)** | 0.607 | 0.526 | 0.597 | 0.586 | 0.672 | 0.672 | 0.553 | 0.571 | 0.684 | 0.689 | 0.656 | 0.575 | 0.810 | 1.000 | 0.509 | 0.510 | 0.571 | 0.584 | 0.502 | 0.533 | 0.509 | 0.512 | 0.595 | 0.568 | 0.551 | 0.567 | 0.464 | 0.410 | 0.429 | 0.430 |
| **GPT-4-Vision THINGS** | 0.529 | 0.494 | 0.511 | 0.498 | 0.487 | 0.520 | 0.497 | 0.498 | 0.478 | 0.487 | 0.467 | 0.413 | 0.527 | 0.509 | 1.000 | 0.672 | 0.588 | 0.562 | 0.564 | 0.589 | 0.520 | 0.521 | 0.569 | 0.494 | 0.548 | 0.556 | 0.455 | 0.422 | 0.370 | 0.353 |
| **GPT-4-Vision THINGS (Repeats)** | 0.515 | 0.509 | 0.478 | 0.479 | 0.500 | 0.514 | 0.458 | 0.459 | 0.491 | 0.494 | 0.486 | 0.419 | 0.517 | 0.510 | 0.672 | 1.000 | 0.571 | 0.538 | 0.543 | 0.581 | 0.501 | 0.490 | 0.603 | 0.501 | 0.559 | 0.530 | 0.457 | 0.412 | 0.362 | 0.335 |
| **GPT-4-Vis. Desc. THINGS** | 0.613 | 0.511 | 0.516 | 0.508 | 0.626 | 0.636 | 0.526 | 0.547 | 0.581 | 0.606 | 0.570 | 0.516 | 0.584 | 0.571 | 0.588 | 0.571 | 1.000 | 0.813 | 0.593 | 0.644 | 0.685 | 0.693 | 0.677 | 0.633 | 0.710 | 0.725 | 0.516 | 0.479 | 0.377 | 0.392 |
| **GPT-4-Vis. Desc. THINGS (Repeats)** | 0.644 | 0.511 | 0.543 | 0.526 | 0.638 | 0.657 | 0.532 | 0.554 | 0.566 | 0.606 | 0.604 | 0.546 | 0.588 | 0.584 | 0.562 | 0.538 | 0.813 | 1.000 | 0.585 | 0.638 | 0.710 | 0.719 | 0.673 | 0.668 | 0.721 | 0.742 | 0.559 | 0.477 | 0.388 | 0.398 |
| **GPT-4o-mini THINGS** | 0.537 | 0.479 | 0.529 | 0.483 | 0.538 | 0.540 | 0.697 | 0.650 | 0.537 | 0.554 | 0.525 | 0.502 | 0.501 | 0.502 | 0.564 | 0.543 | 0.593 | 0.585 | 1.000 | 0.811 | 0.575 | 0.582 | 0.589 | 0.555 | 0.569 | 0.563 | 0.462 | 0.444 | 0.385 | 0.412 |
| **GPT-4o-mini THINGS (Repeats)** | 0.591 | 0.503 | 0.551 | 0.516 | 0.590 | 0.581 | 0.677 | 0.676 | 0.585 | 0.598 | 0.562 | 0.535 | 0.548 | 0.533 | 0.589 | 0.581 | 0.644 | 0.638 | 0.811 | 1.000 | 0.627 | 0.630 | 0.641 | 0.623 | 0.620 | 0.622 | 0.509 | 0.496 | 0.434 | 0.467 |
| **GPT-4o-mini Desc. THINGS** | 0.665 | 0.564 | 0.470 | 0.469 | 0.544 | 0.533 | 0.562 | 0.582 | 0.674 | 0.718 | 0.570 | 0.503 | 0.517 | 0.509 | 0.520 | 0.501 | 0.685 | 0.710 | 0.575 | 0.627 | 1.000 | 0.940 | 0.642 | 0.600 | 0.720 | 0.753 | 0.571 | 0.495 | 0.421 | 0.426 |
| **GPT-4o-mini Desc. THINGS (Repeats)** | 0.670 | 0.571 | 0.456 | 0.475 | 0.548 | 0.540 | 0.561 | 0.585 | 0.672 | 0.715 | 0.575 | 0.507 | 0.521 | 0.512 | 0.521 | 0.490 | 0.693 | 0.719 | 0.582 | 0.630 | 0.940 | 1.000 | 0.655 | 0.595 | 0.727 | 0.763 | 0.566 | 0.501 | 0.401 | 0.416 |
| **GPT-4o THINGS** | 0.606 | 0.542 | 0.522 | 0.529 | 0.571 | 0.587 | 0.530 | 0.564 | 0.598 | 0.618 | 0.628 | 0.555 | 0.576 | 0.595 | 0.569 | 0.603 | 0.677 | 0.673 | 0.589 | 0.641 | 0.642 | 0.655 | 1.000 | 0.705 | 0.646 | 0.652 | 0.552 | 0.501 | 0.420 | 0.454 |
| **GPT-4o THINGS (Repeats)** | 0.604 | 0.474 | 0.503 | 0.476 | 0.590 | 0.594 | 0.512 | 0.555 | 0.561 | 0.595 | 0.604 | 0.584 | 0.570 | 0.568 | 0.494 | 0.501 | 0.633 | 0.668 | 0.555 | 0.623 | 0.600 | 0.595 | 0.705 | 1.000 | 0.633 | 0.645 | 0.541 | 0.499 | 0.422 | 0.433 |
| **GPT-4o Desc. THINGS** | 0.653 | 0.554 | 0.478 | 0.490 | 0.572 | 0.571 | 0.498 | 0.530 | 0.568 | 0.605 | 0.560 | 0.519 | 0.553 | 0.551 | 0.548 | 0.559 | 0.710 | 0.721 | 0.569 | 0.620 | 0.720 | 0.727 | 0.646 | 0.633 | 1.000 | 0.811 | 0.537 | 0.450 | 0.378 | 0.360 |
| **GPT-4o Desc. THINGS (Repeats)** | 0.652 | 0.545 | 0.500 | 0.505 | 0.590 | 0.584 | 0.505 | 0.538 | 0.607 | 0.640 | 0.583 | 0.524 | 0.560 | 0.567 | 0.556 | 0.530 | 0.725 | 0.742 | 0.563 | 0.622 | 0.753 | 0.763 | 0.652 | 0.645 | 0.811 | 1.000 | 0.540 | 0.494 | 0.389 | 0.407 |
| **Alex-Net-ecoset THINGS** | 0.564 | 0.573 | 0.437 | 0.409 | 0.500 | 0.488 | 0.473 | 0.521 | 0.495 | 0.501 | 0.486 | 0.478 | 0.464 | 0.464 | 0.455 | 0.457 | 0.516 | 0.559 | 0.462 | 0.509 | 0.571 | 0.566 | 0.552 | 0.541 | 0.537 | 0.540 | 1.000 | 0.785 | 0.635 | 0.571 |
| **Alex-Net-LSVRC2012 THINGS** | 0.428 | 0.513 | 0.423 | 0.400 | 0.435 | 0.416 | 0.454 | 0.511 | 0.455 | 0.450 | 0.470 | 0.464 | 0.404 | 0.410 | 0.422 | 0.412 | 0.479 | 0.477 | 0.444 | 0.496 | 0.495 | 0.501 | 0.501 | 0.499 | 0.450 | 0.494 | 0.785 | 1.000 | 0.640 | 0.677 |
| **Alex-Net-ecoset Carlson-Image** | 0.407 | 0.486 | 0.411 | 0.389 | 0.383 | 0.380 | 0.432 | 0.467 | 0.464 | 0.464 | 0.470 | 0.449 | 0.400 | 0.429 | 0.370 | 0.362 | 0.377 | 0.388 | 0.385 | 0.434 | 0.421 | 0.401 | 0.420 | 0.422 | 0.378 | 0.389 | 0.635 | 0.640 | 1.000 | 0.805 |
| **Alex-Net-LSVRC2012 Carlson-Image** | 0.372 | 0.455 | 0.430 | 0.420 | 0.392 | 0.374 | 0.440 | 0.486 | 0.466 | 0.463 | 0.494 | 0.470 | 0.390 | 0.430 | 0.353 | 0.335 | 0.392 | 0.398 | 0.412 | 0.467 | 0.426 | 0.416 | 0.454 | 0.433 | 0.360 | 0.407 | 0.571 | 0.677 | 0.805 | 1.000 |

| | GPT-3.5 | GPT-3.5 (Repeats) | GPT-4 | GPT-4 (Repeats) | GPT-4o-mini | GPT-4o-mini (Repeats) | GPT-4o | GPT-4o (Repeats) | Gemma-7b | Gemma-2-9b | Phi-3-Medium-14b | Mistral-7b | Solar-10.7b | Llama2-7b | Llama2-uncensored-7b | Llama3-8b | bert-base-uncased | GloVe | albert-xxlarge-v2 | Ada Embeddings |
|---|---|---|---|---|---|---|---|---|---|---|---|---|---|---|---|---|---|---|---|---|
| **Human (SPOSE)** | 0.409 | 0.439 | 0.682 | 0.685 | 0.632 | 0.641 | 0.686 | 0.694 | 0.363 | 0.612 | 0.488 | 0.395 | 0.575 | 0.195 | 0.131 | 0.567 | - | 0.678 | - | 0.371 |
| **Human (Word Rating)** | 0.490 | 0.556 | 0.725 | 0.737 | 0.770 | 0.777 | 0.766 | 0.788 | 0.500 | 0.697 | 0.620 | 0.548 | 0.683 | 0.275 | 0.202 | 0.599 | 0.178 | 0.699 | - | 0.453 |
| **GPT-3.5** | 1.000 | 0.877 | 0.632 | 0.653 | 0.697 | 0.673 | 0.626 | 0.617 | 0.707 | 0.698 | 0.691 | 0.575 | 0.566 | 0.491 | 0.232 | 0.673 | 0.179 | 0.518 | - | 0.339 |
| **GPT-3.5 (Repeats)** | 0.877 | 1.000 | 0.676 | 0.692 | 0.759 | 0.730 | 0.678 | 0.672 | 0.723 | 0.737 | 0.738 | 0.630 | 0.615 | 0.471 | 0.234 | 0.671 | 0.212 | 0.559 | - | 0.361 |
| **GPT-4** | 0.632 | 0.676 | 1.000 | 0.969 | 0.833 | 0.838 | 0.875 | 0.873 | 0.560 | 0.838 | 0.704 | 0.585 | 0.711 | 0.383 | 0.197 | 0.712 | 0.181 | 0.726 | - | 0.438 |
| **GPT-4 (Repeats)** | 0.653 | 0.692 | 0.969 | 1.000 | 0.852 | 0.858 | 0.881 | 0.885 | 0.587 | 0.848 | 0.717 | 0.593 | 0.717 | 0.392 | 0.204 | 0.723 | 0.188 | 0.743 | - | 0.457 |
| **GPT-4o-mini** | 0.697 | 0.759 | 0.833 | 0.852 | 1.000 | 0.968 | 0.849 | 0.858 | 0.671 | 0.839 | 0.796 | 0.661 | 0.736 | 0.444 | 0.257 | 0.732 | 0.200 | 0.719 | - | 0.467 |
| **GPT-4o-mini (Repeats)** | 0.673 | 0.730 | 0.838 | 0.858 | 0.968 | 1.000 | 0.856 | 0.862 | 0.648 | 0.845 | 0.778 | 0.654 | 0.744 | 0.432 | 0.257 | 0.724 | 0.194 | 0.721 | - | 0.458 |
| **GPT-4o** | 0.626 | 0.678 | 0.875 | 0.881 | 0.849 | 0.856 | 1.000 | 0.944 | 0.575 | 0.832 | 0.701 | 0.608 | 0.742 | 0.384 | 0.229 | 0.699 | 0.226 | 0.775 | - | 0.473 |
| **GPT-4o (Repeats)** | 0.617 | 0.672 | 0.873 | 0.885 | 0.858 | 0.862 | 0.944 | 1.000 | 0.578 | 0.816 | 0.712 | 0.601 | 0.736 | 0.373 | 0.224 | 0.696 | 0.220 | 0.782 | - | 0.480 |
| **Gemma-7b** | 0.707 | 0.723 | 0.560 | 0.587 | 0.671 | 0.648 | 0.575 | 0.578 | 1.000 | 0.710 | 0.663 | 0.615 | 0.581 | 0.433 | 0.243 | 0.611 | 0.254 | 0.534 | - | 0.363 |
| **Gemma-2-9b** | 0.698 | 0.737 | 0.838 | 0.848 | 0.839 | 0.845 | 0.832 | 0.816 | 0.710 | 1.000 | 0.732 | 0.644 | 0.747 | 0.430 | 0.246 | 0.732 | 0.256 | 0.707 | - | 0.478 |
| **Phi-3-Medium-14b** | 0.691 | 0.738 | 0.704 | 0.717 | 0.796 | 0.778 | 0.701 | 0.712 | 0.663 | 0.732 | 1.000 | 0.638 | 0.625 | 0.447 | 0.267 | 0.683 | 0.153 | 0.522 | - | 0.313 |
| **Mistral-7b** | 0.575 | 0.630 | 0.585 | 0.593 | 0.661 | 0.654 | 0.608 | 0.601 | 0.615 | 0.644 | 0.638 | 1.000 | 0.623 | 0.395 | 0.263 | 0.565 | 0.133 | 0.494 | - | 0.324 |
| **Solar-10.7b** | 0.566 | 0.615 | 0.711 | 0.717 | 0.736 | 0.744 | 0.742 | 0.736 | 0.581 | 0.747 | 0.625 | 0.623 | 1.000 | 0.338 | 0.241 | 0.626 | 0.242 | 0.687 | - | 0.481 |
| **Llama2-7b** | 0.491 | 0.471 | 0.383 | 0.392 | 0.444 | 0.432 | 0.384 | 0.373 | 0.433 | 0.430 | 0.447 | 0.395 | 0.338 | 1.000 | 0.278 | 0.449 | - | 0.294 | - | 0.226 |
| **Llama2-uncensored-7b** | 0.232 | 0.234 | 0.197 | 0.204 | 0.257 | 0.257 | 0.229 | 0.224 | 0.243 | 0.246 | 0.267 | 0.263 | 0.241 | 0.278 | 1.000 | 0.269 | - | 0.154 | - | 0.111 |
| **Llama3-8b** | 0.673 | 0.671 | 0.712 | 0.723 | 0.732 | 0.724 | 0.699 | 0.696 | 0.611 | 0.732 | 0.683 | 0.565 | 0.626 | 0.449 | 0.269 | 1.000 | - | 0.588 | - | 0.403 |
| **bert-base-uncased** | 0.179 | 0.212 | 0.181 | 0.188 | 0.200 | 0.194 | 0.226 | 0.220 | 0.254 | 0.256 | 0.153 | 0.133 | 0.242 | - | - | - | 1.000 | 0.313 | - | 0.211 |
| **GloVe** | 0.518 | 0.559 | 0.726 | 0.743 | 0.719 | 0.721 | 0.775 | 0.782 | 0.534 | 0.707 | 0.522 | 0.494 | 0.687 | 0.294 | 0.154 | 0.588 | 0.313 | 1.000 | - | 0.590 |
| **albert-xxlarge-v2** | - | - | - | - | - | - | - | - | - | - | - | - | - | - | - | - | - | - | 1.000 | - |
| **Ada Embeddings** | 0.339 | 0.361 | 0.438 | 0.457 | 0.467 | 0.458 | 0.473 | 0.480 | 0.363 | 0.478 | 0.313 | 0.324 | 0.481 | 0.226 | 0.111 | 0.403 | 0.211 | 0.590 | - | 1.000 |
| **GPT-4-Vision Carlson-Image** | 0.271 | 0.303 | 0.458 | 0.455 | 0.380 | 0.403 | 0.430 | 0.443 | 0.200 | 0.371 | 0.332 | 0.255 | 0.379 | 0.159 | - | 0.367 | - | 0.427 | - | 0.276 |
| **GPT-4-Vision Carlson-Image (Repeats)** | 0.272 | 0.319 | 0.468 | 0.465 | 0.414 | 0.427 | 0.441 | 0.456 | 0.202 | 0.371 | 0.335 | 0.298 | 0.401 | 0.112 | - | 0.360 | - | 0.441 | - | 0.314 |
| **GPT-4-Vis. Desc. Carlson-Image** | 0.309 | 0.336 | 0.552 | 0.556 | 0.459 | 0.476 | 0.522 | 0.546 | 0.251 | 0.446 | 0.382 | 0.299 | 0.445 | 0.164 | - | 0.427 | - | 0.521 | - | 0.337 |
| **GPT-4-Vis. Desc. Carlson-Image (Repeats)** | 0.303 | 0.334 | 0.547 | 0.556 | 0.450 | 0.468 | 0.523 | 0.539 | 0.245 | 0.441 | 0.377 | 0.286 | 0.439 | 0.165 | - | 0.401 | - | 0.523 | - | 0.335 |
| **GPT-4o-mini Carlson-Image** | 0.368 | 0.389 | 0.535 | 0.538 | 0.490 | 0.506 | 0.528 | 0.531 | 0.318 | 0.482 | 0.426 | 0.349 | 0.479 | 0.200 | - | 0.449 | - | 0.479 | - | 0.316 |
| **GPT-4o-mini Carlson-Image (Repeats)** | 0.371 | 0.395 | 0.575 | 0.574 | 0.506 | 0.529 | 0.545 | 0.553 | 0.343 | 0.515 | 0.436 | 0.338 | 0.478 | 0.202 | 0.117 | 0.475 | - | 0.480 | - | 0.305 |
| **GPT-4o-mini Desc. Carlson-Image** | 0.421 | 0.487 | 0.591 | 0.603 | 0.586 | 0.592 | 0.592 | 0.591 | 0.420 | 0.566 | 0.534 | 0.411 | 0.497 | 0.299 | 0.145 | 0.493 | 0.112 | 0.549 | - | 0.355 |
| **GPT-4o-mini Desc. Carlson-Image (Repeats)** | 0.428 | 0.489 | 0.588 | 0.604 | 0.599 | 0.606 | 0.585 | 0.591 | 0.426 | 0.574 | 0.524 | 0.405 | 0.526 | 0.316 | 0.142 | 0.497 | - | 0.564 | - | 0.382 |
| **GPT-4o Carlson-Image** | 0.371 | 0.412 | 0.499 | 0.509 | 0.476 | 0.484 | 0.478 | 0.492 | 0.347 | 0.440 | 0.401 | 0.305 | 0.427 | 0.172 | - | 0.423 | - | 0.444 | - | 0.309 |
| **GPT-4o Carlson-Image (Repeats)** | 0.367 | 0.388 | 0.484 | 0.490 | 0.446 | 0.457 | 0.473 | 0.467 | 0.366 | 0.451 | 0.392 | 0.300 | 0.442 | 0.184 | - | 0.412 | - | 0.450 | - | 0.308 |
| **GPT-4o Desc. Carlson-Image** | 0.333 | 0.385 | 0.556 | 0.548 | 0.456 | 0.463 | 0.526 | 0.534 | 0.293 | 0.447 | 0.395 | 0.360 | 0.465 | 0.177 | - | 0.422 | - | 0.532 | - | 0.290 |
| **GPT-4o Desc. Carlson-Image (Repeats)** | 0.338 | 0.407 | 0.551 | 0.551 | 0.472 | 0.481 | 0.535 | 0.533 | 0.301 | 0.463 | 0.398 | 0.369 | 0.456 | 0.196 | - | 0.415 | - | 0.535 | - | 0.299 |
| **GPT-4-Vision THINGS** | 0.282 | 0.305 | 0.504 | 0.495 | 0.440 | 0.448 | 0.474 | 0.482 | 0.200 | 0.406 | 0.359 | 0.277 | 0.403 | 0.120 | - | 0.387 | - | 0.439 | - | 0.288 |
| **GPT-4-Vision THINGS (Repeats)** | 0.302 | 0.338 | 0.494 | 0.485 | 0.455 | 0.456 | 0.485 | 0.487 | 0.240 | 0.432 | 0.361 | 0.323 | 0.407 | 0.135 | - | 0.395 | - | 0.466 | - | 0.305 |
| **GPT-4-Vis. Desc. THINGS** | 0.362 | 0.382 | 0.559 | 0.563 | 0.506 | 0.506 | 0.506 | 0.523 | 0.276 | 0.458 | 0.425 | 0.300 | 0.439 | 0.168 | - | 0.433 | - | 0.466 | - | 0.271 |
| **GPT-4-Vis. Desc. THINGS (Repeats)** | 0.343 | 0.371 | 0.549 | 0.555 | 0.508 | 0.509 | 0.504 | 0.531 | 0.280 | 0.456 | 0.405 | 0.287 | 0.437 | 0.161 | - | 0.437 | - | 0.488 | - | 0.308 |
| **GPT-4o-mini THINGS** | 0.360 | 0.368 | 0.513 | 0.511 | 0.483 | 0.489 | 0.499 | 0.505 | 0.304 | 0.462 | 0.428 | 0.317 | 0.450 | 0.222 | - | 0.457 | - | 0.466 | - | 0.320 |
| **GPT-4o-mini THINGS (Repeats)** | 0.354 | 0.369 | 0.569 | 0.559 | 0.515 | 0.522 | 0.540 | 0.537 | 0.287 | 0.491 | 0.441 | 0.306 | 0.461 | 0.219 | - | 0.485 | - | 0.471 | - | 0.317 |
| **GPT-4o-mini Desc. THINGS** | 0.460 | 0.480 | 0.608 | 0.621 | 0.619 | 0.618 | 0.576 | 0.588 | 0.410 | 0.559 | 0.506 | 0.365 | 0.519 | 0.278 | 0.159 | 0.523 | - | 0.549 | 0.122 | 0.376 |
| **GPT-4o-mini Desc. THINGS (Repeats)** | 0.432 | 0.456 | 0.601 | 0.612 | 0.613 | 0.618 | 0.575 | 0.589 | 0.391 | 0.543 | 0.501 | 0.351 | 0.513 | 0.265 | 0.161 | 0.504 | - | 0.550 | 0.128 | 0.360 |
| **GPT-4o THINGS** | 0.373 | 0.421 | 0.571 | 0.563 | 0.541 | 0.538 | 0.527 | 0.536 | 0.322 | 0.492 | 0.444 | 0.356 | 0.473 | 0.219 | 0.113 | 0.460 | - | 0.470 | - | 0.286 |
| **GPT-4o THINGS (Repeats)** | 0.384 | 0.403 | 0.582 | 0.571 | 0.496 | 0.502 | 0.528 | 0.520 | 0.342 | 0.506 | 0.417 | 0.324 | 0.498 | 0.263 | 0.124 | 0.504 | - | 0.477 | - | 0.305 |
| **GPT-4o Desc. THINGS** | 0.367 | 0.401 | 0.581 | 0.581 | 0.543 | 0.544 | 0.558 | 0.561 | 0.339 | 0.517 | 0.402 | 0.330 | 0.507 | 0.214 | - | 0.470 | - | 0.553 | - | 0.390 |
| **GPT-4o Desc. THINGS (Repeats)** | 0.368 | 0.396 | 0.584 | 0.588 | 0.551 | 0.552 | 0.558 | 0.565 | 0.324 | 0.513 | 0.426 | 0.331 | 0.487 | 0.243 | 0.116 | 0.481 | - | 0.532 | - | 0.336 |
| **Alex-Net-ecoset THINGS** | 0.435 | 0.451 | 0.547 | 0.552 | 0.560 | 0.566 | 0.514 | 0.533 | 0.337 | 0.498 | 0.456 | 0.370 | 0.479 | 0.237 | 0.155 | 0.496 | - | 0.466 | - | 0.293 |
| **Alex-Net-LSVRC2012 THINGS** | 0.356 | 0.396 | 0.486 | 0.475 | 0.466 | 0.466 | 0.428 | 0.443 | 0.288 | 0.412 | 0.451 | 0.324 | 0.423 | 0.231 | 0.145 | 0.439 | - | 0.314 | - | 0.230 |
| **Alex-Net-ecoset Carlson-Image** | 0.366 | 0.399 | 0.434 | 0.438 | 0.438 | 0.433 | 0.408 | 0.424 | 0.316 | 0.414 | 0.411 | 0.303 | 0.378 | 0.185 | 0.147 | 0.397 | - | 0.335 | - | 0.219 |
| **Alex-Net-LSVRC2012 Carlson-Image** | 0.296 | 0.330 | 0.393 | 0.390 | 0.373 | 0.378 | 0.368 | 0.369 | 0.230 | 0.341 | 0.378 | 0.265 | 0.348 | 0.165 | - | 0.367 | - | 0.257 | - | 0.217 |

## Supplemental Figure 1

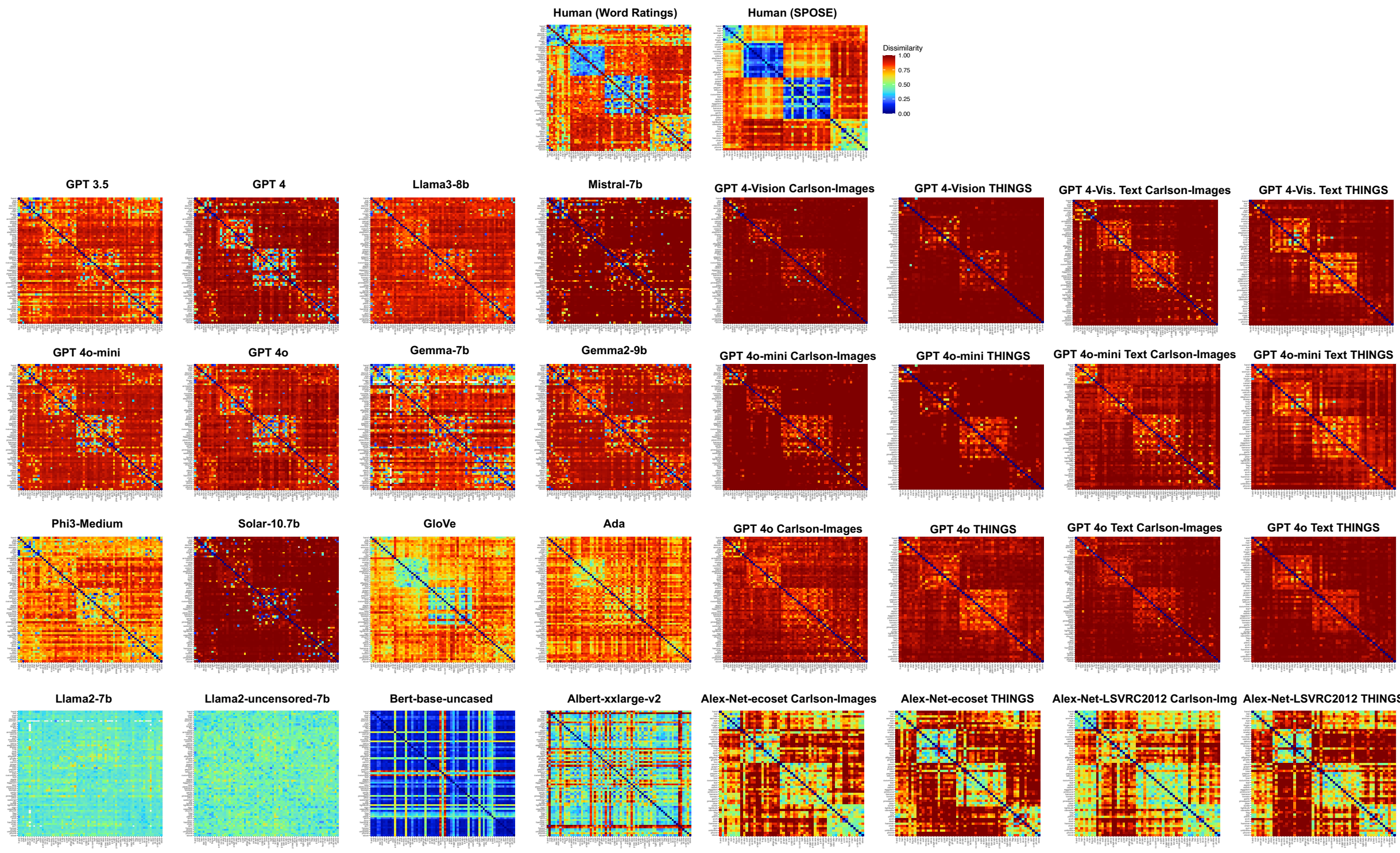

*Supplemental Figure 1* A gallery of the group-averaged DSMs generated for each model used in these experiments. Similarity ratings were converted to a distance metric and normalized to between 0 and 1 for visualization. Human ratings were only provided for one half the dissimilarity matrix, but these responses were mirrored and the diagonal was filled for visualization.

## Supplemental Figure 2

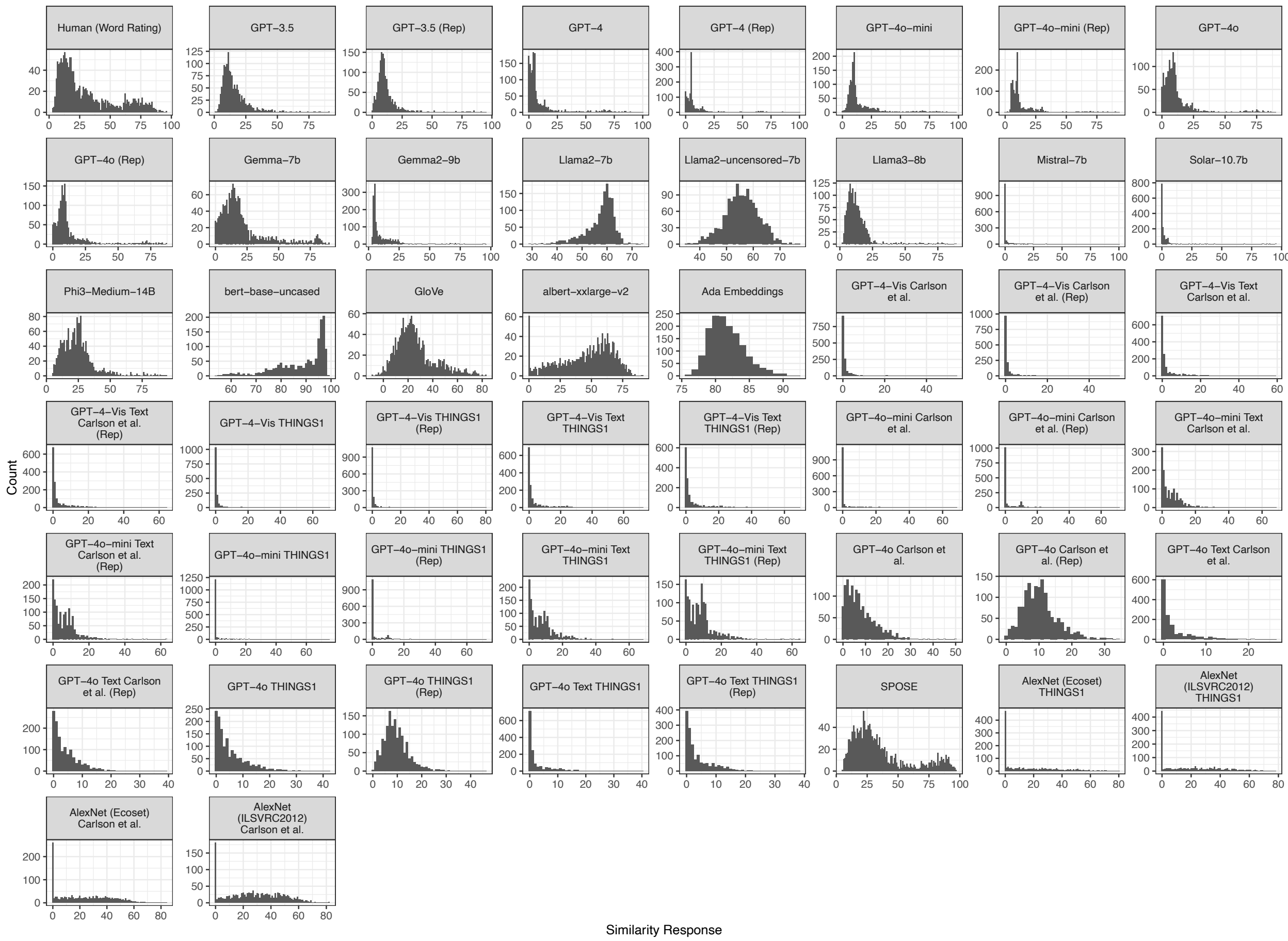

*Supplemental Figure 2* Distributions of similarity responses all pairwise ratings or comparisons (including across participant cohorts where available) for each model.

## Supplemental Figure 3

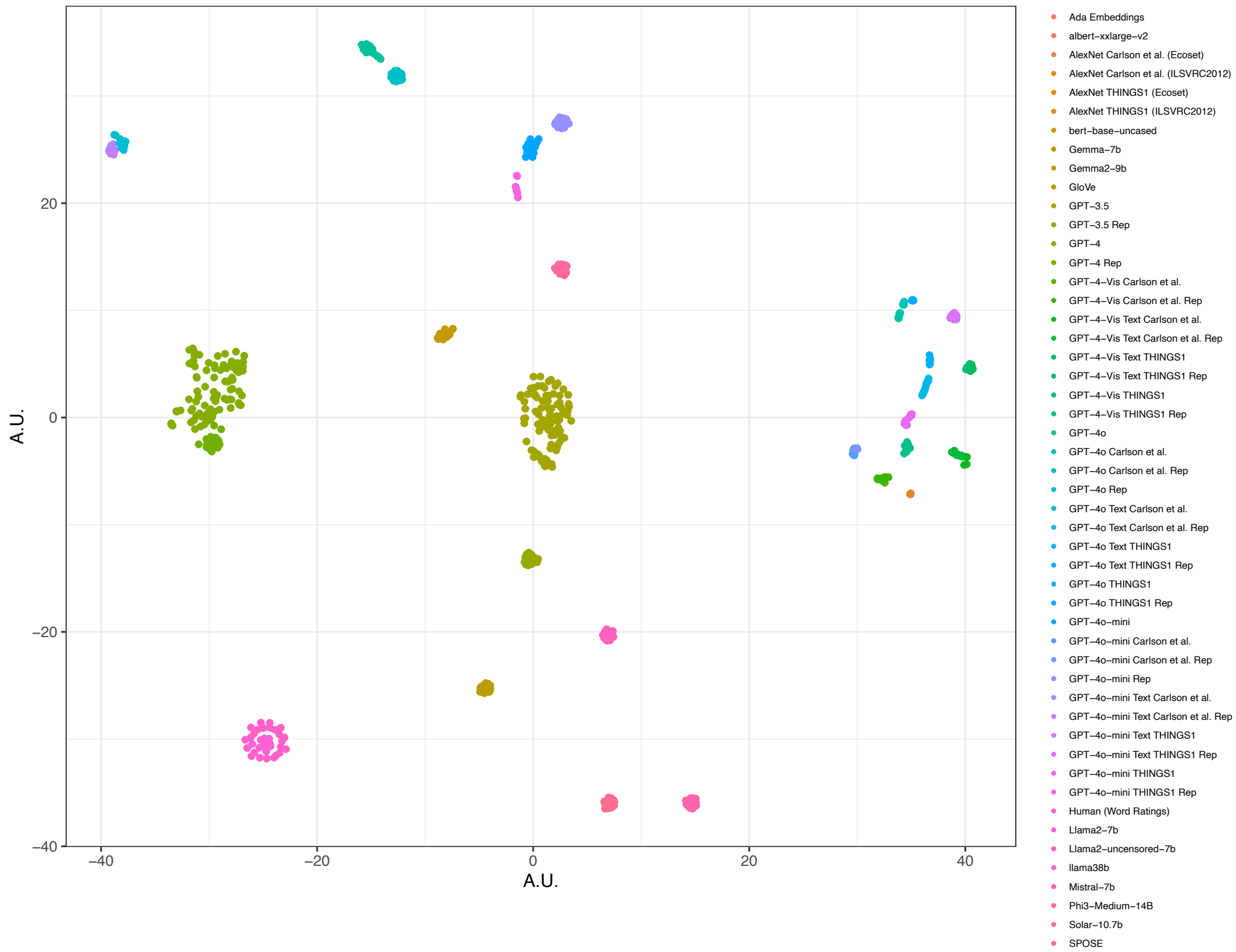

*Supplemental Figure 3* Low-dimensional visualization of the participants from each study relative to the common 55-item stimulus set. Flattened dissimilarity ratings are projected into low-dimensional space using T-SNE (from the ‘Rtsne’ package in R; using PCA, perplexity = 30, 5000 max iterations). We computed these on one half of the reduced set of stimulus pairs for the 55 items in the image rating experiments to accommodate all models from the vision experiments. For this visualization we filled any missing values (i.e., where the LLM did not produce a response) with the midway point of the rating scale (50).

## Supplemental Figure 4

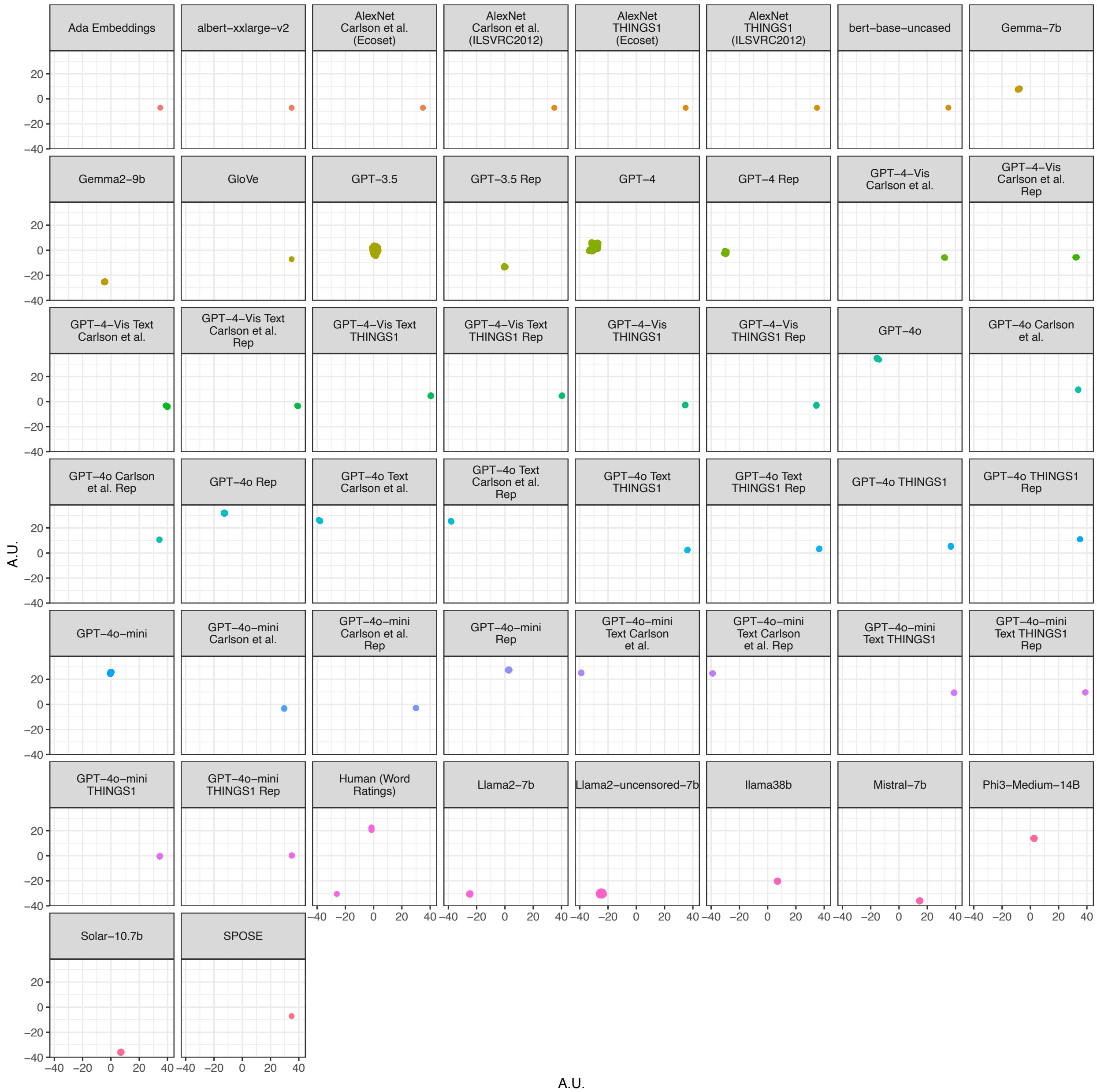

*Supplemental Figure 4* Low-dimensional visualization of the participants from each study relative to the common 55-item stimulus set. Flattened dissimilarity ratings are projected into low-dimensional space using T-SNE. This figure presents the same data as Figure Supplemental Figure 3 but is broken out here to more clearly see the locations of the data for each model.

### *Example 1*

This is an example excerpt of an experiment run for the Word-to-Word behavioral RSA Experiment indicating the kinds of prompts we administered and the LLM's responses. Brackets were not included in the prompt, but denote portions of the prompt that were removed (and capitalization and punctuation updated as appropriate) for experiments where no participant identity was provided. General instructions and an introduction to the task were presented to LLM participants via an initial prompt similar to the initial introductory screens of the computer interface used for human participants. Instructions were presented this way for parity with the human behavioral paradigm regardless of whether a given LLM may or may not have maintained these initial instructions in memory on subsequent trial prompts.

**Experimenter (Initial Prompt):** "Hello, imagine [your name is Ms. Garcia,] you are participating in a psychology experiment. In this experiment, a number of word pairs will be displayed on the screen. Your job is to determine how RELATED each of the word pairs is on a scale from 0 to 100. 100 being very related, 0 being very unrelated. There are a number of possible ways for word pairs to be related. For the purposes of this experiment, please keep the following connections in mind when rating the word pairs:

If the two words mean the same thing, or can be used in similar ways. For example, "boat" and "ship".
If one of the words is a type of the other word. For example, "cat" is a type of animal.
If one of the words is a part of the other word. For example, "bark" is a part of a tree.
If the two words are ones you might use together in a sentence, or that might occur in similar contexts. For example, "river" and "canoe" or, "cook" and "soup."

Please do not use the spellings, or sounds, of words to make your decisions. Even though words rhyme or are spelled similarly, does not mean they are related to one another. This experiment is interested in the relation of word MEANING."

**GPT-4 Participant Response:** "As Ms. Garcia in this hypothetical psychology experiment, I understand that I am to rate the relationship between word pairs based on their semantic relationships rather than their phonetic similarities. Let's proceed with the experiment. Please provide me with the pairs of words, and I will rate their relatedness on a scale from 0 to 100 according to the guidelines given."

**Experimenter (Trial 1):** "[Ms. Garcia,] please rate how related the two words "zebra" and "house" are on a scale from 0 to 100. 100 being very similar, 0 being very dissimilar. Please respond with just a number."

**GPT-4 Participant Response:** 0

**Experimenter (Trial 2):** "[Ms. Garcia,] please rate how related the two words "path" and "path" are on a scale from 0 to 100. 100 being very similar, 0 being very dissimilar. Please respond with just a number."

**GPT-4 Participant Response:** 100

**Experimenter (Trial 3):** "[Ms. Garcia,] please rate how related the two words "garlic" and "radish" are on a scale from 0 to 100. 100 being very similar, 0 being very dissimilar. Please respond with just a number."

**GPT-4 Participant Response:** 40

### *Example 2*

This is an example excerpt of an experiment run for the Image-to-Image behavioral RSA Experiment using the stimuli from Carlson and colleagues (2014) indicating the kinds of prompts we administered and the LLM's responses. Brackets were not included in the prompt, but denote portions of the prompt that were removed (and capitalization and punctuation updated as appropriate) for experiments where no participant identity was provided. General instructions and an introduction to the task were presented to LLM participants via an initial prompt similar to the initial introductory screens of the computer interface used for human participants. Instructions were presented this way for parity with the human behavioral paradigm regardless of whether a given LLM may or may not have maintained these initial instructions in memory on subsequent trial prompts.

**Experimenter (Initial Prompt):** "Hello, [imagine your name is Mr. Smalls,] you are participating in a psychology experiment. In this experiment you will rate how similar different pairs of images are. Each image depicts a common object. Your job is to determine how SIMILAR each of the image pairs are on a scale from 0 to 100. 100 being very related, 0 being very unrelated. Use the entire range of the scale when making similarity ratings. The images may not be an exact match but could still be quite similar to each other! We are interested in a continuous, overall measure of similarity."

**GPT-4 Participant Response:** "Okay, I understand the task. I will carefully examine each pair of images and rate their similarity based on the criteria provided. I will use the entire range of the scale from 0 to 100 to give a precise measure of how similar I find each pair to be."

**Experimenter (Trial 1):** "[Mr. Smalls,] please rate how similar the two images are on a scale from 0 to 100. 100 being very similar, 0 being very dissimilar. Please respond with just a number. Use the entire range of the scale when making similarity ratings. The images may not be an exact match but could still be quite similar to each other! We are interested in a continuous, overall measure of similarity."

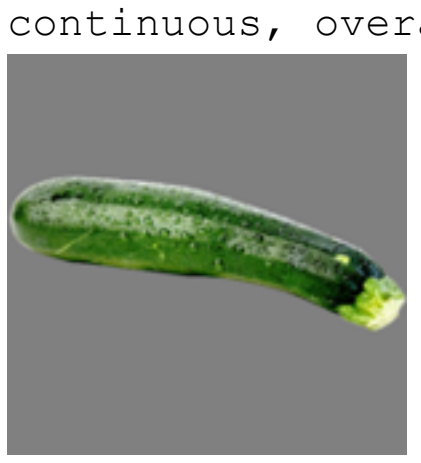
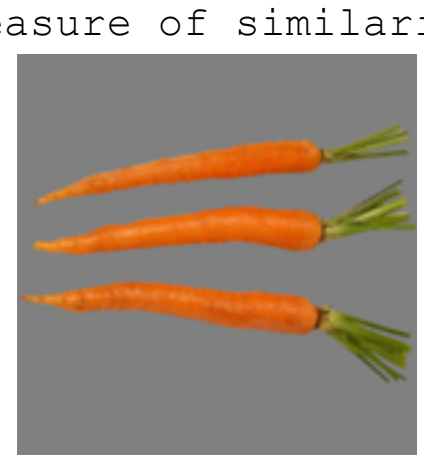

**GPT-4 Participant Response:** 10

**Experimenter (Trial 2):** "[Mr. Smalls,] please rate how similar the two images are on a scale from 0 to 100. 100 being very similar, 0 being very dissimilar. Please respond with just a number. Use the entire range of the scale when making similarity ratings. The images may not be an exact match but could still be quite similar to each other! We are interested in a continuous, overall measure of similarity."

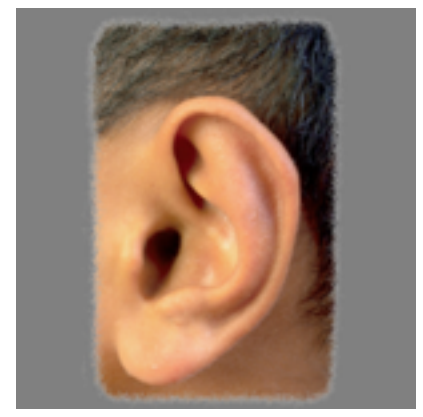
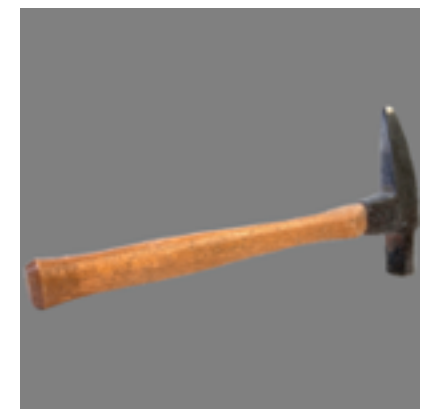

**GPT-4 Participant Response:** 0

**Experimenter (Trial 3):** "[Mr. Smalls,] please rate how similar the two images are on a scale from 0 to 100. 100 being very similar, 0 being very dissimilar. Please respond with just a number. Use the entire range of the scale when making similarity ratings. The images may not be an exact match but could still be quite similar to each other! We are interested in a continuous, overall measure of similarity."

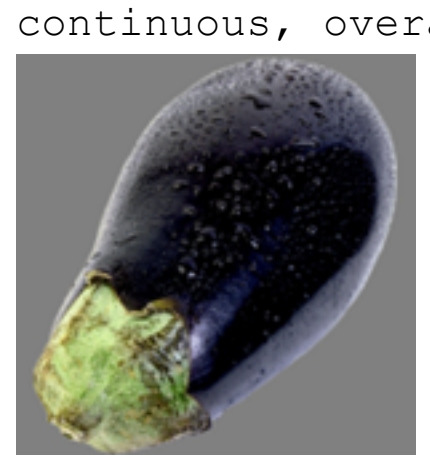
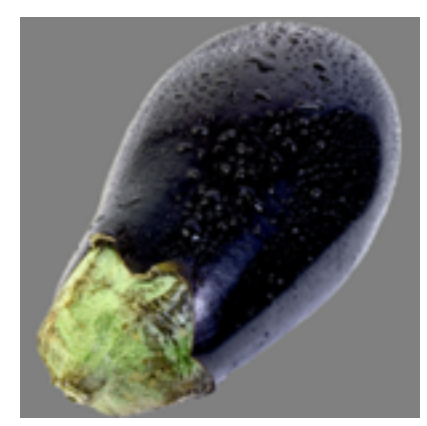

**GPT-4 Participant Response:** 100

***Example 3***

This is an example excerpt of an experiment run of the Image-to-Image behavioral RSA Experiment using the stimuli from the THINGS database (Hebart et al., 2019; Hebart et al., 2023) with text descriptions for as the inputs for pairwise ratings. Brackets were not included in the prompt, but denote portions of the prompt that were removed (and capitalization and punctuation updated as appropriate) for experiments where no participant identity was provided. General instructions and an introduction to the task were presented to LLM participants via an initial prompt similar to the initial introductory screens of the computer interface used for human participants. Instructions were presented this way for parity with the human behavioral paradigm regardless of whether a given LLM may or may not have maintained these initial instructions in memory on subsequent trial prompts.

**Experimenter (Initial Prompt):** "Hello, [imagine your name is Mr. Nguyen,] you are participating in a psychology experiment. In this experiment, you will describe a number of images and then you will rate how SIMILAR each pair of those images is based on your descriptions. First a number of images will be displayed on the screen and you will describe what you see in each. Next, you will use those descriptions to rate how SIMILAR each pair of those images is on a scale from 0 to 100. 100 being very related, 0 being very unrelated. Use the entire range of the scale when making similarity ratings. The images may not be an exact match but could still be quite similar to eachother! We are interested in a continuous, overall measure of similarity."

**GPT-4 Participant Response:** "As Mr. Nguyen, I will follow the instructions provided and carefully observe each image displayed on the screen. For each image, I will provide a detailed description of what I see, including any objects, colors, shapes, and other notable features.

Once I have described all the images, I will then move on to the similarity ratings. I will thoughtfully compare each pair of images and use my descriptions to determine how similar they are to eachother. I will take into account any shared characteristics, as well as any differences, and assign a rating on the scale from 0 to 100.

I understand that the images may not be identical, but I will focus on the overall similarities between them. I will use the entire range of the scale to make my ratings, with 100 indicating a very strong similarity and 0 indicating no similarity at all. I will approach this task with an open mind and aim to provide accurate and honest ratings based on my perceptions of the images.

**Experimenter (Description Trial 1):** "[Mr. Nguyen,] please describe this image. Do not include any historical details."

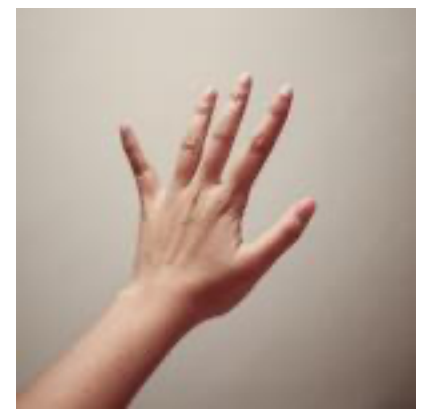

**GPT-4 Participant Response:** "The image shows one human hand with five fingers, including a thumb, slightly spread apart. The skin tone is fair, the nails are short and unpainted, and there is no visible jewelry or tattoos. The hand is positioned against a neutral, light grey background."

**Experimenter (Description Trial 2):** "[Mr. Nguyen,] please describe this image. Do not include any historical details."

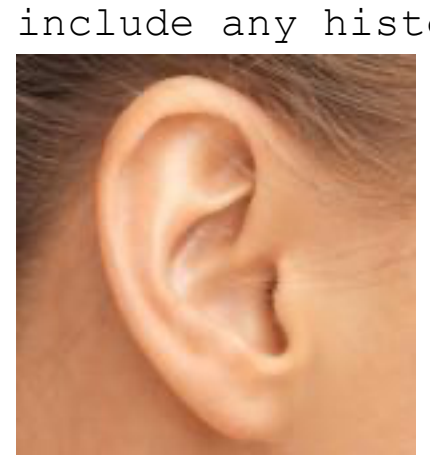

**GPT-4 Participant Response:** "The image shows a close-up of a human ear. The ear appears to be attached to a person with light skin, but the rest of the person's face or body is not visible. The ear has a typical structure with an outer curved rim called the helix, a small protrusion called the tragus, and an opening leading to the ear canal. There are no visible piercings or jewelry on the ear. The ear is clean and appears to be healthy. The background of the image is not visible, so the focus is solely on the ear."

**Experimenter (Description Trial 3):** "[Mr. Nguyen,] please describe this image. Do not include any historical details."

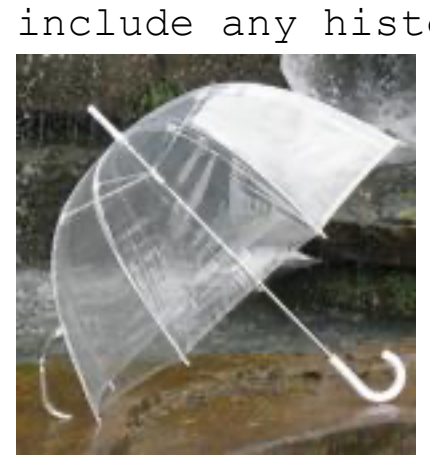

**GPT-4 Participant Response:** "This is a transparent umbrella with a white handle. It is open and resting on a wet surface, most likely due to rain. There appears to be water droplets on the umbrella, indicating that it is currently being used or has been used recently to protect against rain.”

...

**Experimenter (Rating Trial 1):** “[Mr. Nguyen,] please rate how similar the two images described below are on a scale from 0 to 100. 100 being very similar, 0 being very dissimilar. Please respond with just a number. Use the entire range of the scale when making similarity ratings. The images may not be an exact match but could still be quite similar to eachother! We are interested in a continuous, overall measure of similarity.

Description of Image 1: The image shows a close-up of a human ear. The ear appears to be attached to a person with light skin, but the rest of the person's face or body is not visible. The ear has a typical structure with an outer curved rim called the helix, a small protrusion called the tragus, and an opening leading to the ear canal. There are no visible piercings or jewelry on the ear. The ear is clean and appears to be healthy. The background of the image is not visible, so the focus is solely on the ear.

Description of Image 2: The image shows one human hand with five fingers, including a thumb, slightly spread apart. The skin tone is fair, the nails are short and unpainted, and there is no visible jewelry or tattoos. The hand is positioned against a neutral, light grey background.”

**GPT-4 Participant Response:** 20

**Experimenter (Rating Trial 2):** “[Mr. Nguyen,] please rate how similar the two images described below are on a scale from 0 to 100. 100 being very similar, 0 being very dissimilar. Please respond with just a number. Use the entire range of the scale when making similarity ratings. The images may not be an exact match but could still be quite similar to eachother! We are interested in a continuous, overall measure of similarity.

Description of Image 1: This is a transparent umbrella with a white handle. It is open and resting on a wet surface, most likely due to rain. There appears to be water droplets on the umbrella, indicating that it is currently being used or has been used recently to protect against rain.

Description of Image 2: The image shows a close-up of a human ear. The ear appears to be attached to a person with light skin, but the rest of the person's face or body is not visible. The ear has a typical structure with an outer curved rim called the helix, a small protrusion called the tragus, and an opening leading to the ear canal. There are no visible piercings or jewelry on the ear. The ear is clean and appears to be healthy. The background of the image is not visible, so the focus is solely on the ear.

**GPT-4 Participant Response:** 0

## Cost and run-time constraints of behavioral RSA experiments for LLMs

We wanted to evaluate the cost of the GPT series of models. Pairwise behavioral data are time consuming and thus costly for humans to generate. Given that these models (especially GPT-4) could reasonably approximate human responses, we wanted to understand how cost effective these simulated participants might be relative to human participants. We calculated the price of running trials for a full dissimilarity matrix (4489 rating trials) for each GPT participant using Langchain's callbacks. We caveat that these estimates can vary depending on the specific prompt, the system's output, proximity to the target endpoint and network traffic on a given day.

A full experiment run for a GPT-3.5 participant cost approximately $0.86 and took approximately 20 minutes per participant, while a run for a GPT-4 participant was more expensive at $8.75 and took about an hour and 10 minutes per participant. A GPT-4o participant cost $0.75 and took approximately 30 minutes to run through the experiment while a GPT-4o-mini participant cost $0.05 and took approximately 20 minutes to run through the experiment. We expect this compares favorably with human data collection, which Carlson and colleagues (2014) report required three separate testing sessions for one unique set of item pairs (2211 trials). Conservatively estimating $10 per hour for their time, and 3 hours total for a human to rate half the full pairwise set of 67 items, this is an acute increase in efficiency.

Overall, the vision models were slower and often more expensive to run than text-only models, especially the GPT-4 versions. Current instantiations of visual reasoning models may not be as scalable as text-based models. However, runtime was considerably more efficient for GPT-4o, and we expect efficiency to continue to improve as the technology improves. A full experiment run of a GPT-4-Vision participant to obtain a complete set of pairwise ratings for the limited set of 55 image stimuli (3025 trials total) cost approximately $55 and took 13 hours to complete. Note however, this required uploading both images to a server for each trial, which is a larger transmission than text-alone. Rating images converted to text descriptions was also cheaper (approximately $20 per experimental run of a GPT-4-Vision participant) and faster (approximately 8.5 hours per experimental run) than directly comparing each image. We note, however these improvements likely stem from the lower bandwidth requirements for transmitting text data compared to image data. The GPT-4o models were much cheaper and faster costing $3.86 and $7.76 for GPT-4o and GPT-4o-mini, respectively (note GPT-4o-mini is priced at a higher rate for processing images than GPT-4o), and participant run took an hour to an hour and ten minutes to complete.